\documentclass[conference]{IEEEtran}
\usepackage{cite}

\ifCLASSINFOpdf
  \usepackage[pdftex]{graphicx}
  \graphicspath{{./images/}}
  \DeclareGraphicsExtensions{.eps}
\else
\fi
\usepackage[cmex10]{amsmath}
\usepackage{amssymb}             
\newcommand{\R}{\mathbb{R}}

\usepackage{algorithmic}
\usepackage{algorithm}

\floatstyle{plaintop}               
\restylefloat{algorithm}            

\usepackage[font=footnotesize]{subfig}
\usepackage{booktabs}
\usepackage{makecell} 
\usepackage{tabularx}
\usepackage{color, colortbl} 
\definecolor{Lightgrey}{gray}{0.9}
\definecolor{Darkgrey}{gray}{0.6}
\definecolor{Red}{rgb}{1,0,0}
\definecolor{White}{rgb}{1,1,1}

\newcommand{\dc}{\cellcolor{Darkgrey}}
\newcommand{\rc}{\cellcolor{Red}}

\usepackage{fancyvrb}

\begin{document}
%
\title{ConFoc: Content-Focus Protection Against \\ Trojan Attacks on Neural Networks}

\author{
   \IEEEauthorblockN{Miguel Villarreal-Vasquez and Bharat Bhargava}
  \IEEEauthorblockA{
   Purdue University\\
   Email: \{mvillar,bbshail\}@purdue.edu}
}

\maketitle

\begin{abstract}
Deep Neural Networks (DNNs) have been applied successfully in computer vision. 
However, their wide adoption in image-related applications is threatened by their 
vulnerability to trojan attacks. 
%
These attacks insert some misbehavior at training using samples 
with a mark or trigger, which is exploited at inference or testing time.
In this work, we analyze the composition of the features learned by DNNs at training.
%
We identify that they, including those related to the inserted triggers, contain 
both content (semantic information) and style (texture information),
which are recognized as a whole by DNNs at testing time. 
%
We then propose a novel defensive technique against trojan attacks, in which
DNNs are taught to disregard the styles of inputs and focus on their content only 
to mitigate the effect of triggers during the classification.
%
The generic applicability of the approach is demonstrated in the 
context of a traffic sign and a face recognition application. 
Each of them is exposed to a different attack with a variety of triggers. 
Results show that the method reduces the attack success rate significantly 
to values $< 1\%$ in all the tested attacks while keeping as well as improving the initial 
accuracy of the models when processing both benign and adversarial data.
\end{abstract}

\section{Introduction}\label{sec:introduction}

\begin{figure*}[!t]
\centering
\subfloat[Feature-Focus of Humans and Traditionally Trained DNNs]{\includegraphics[width=3in]{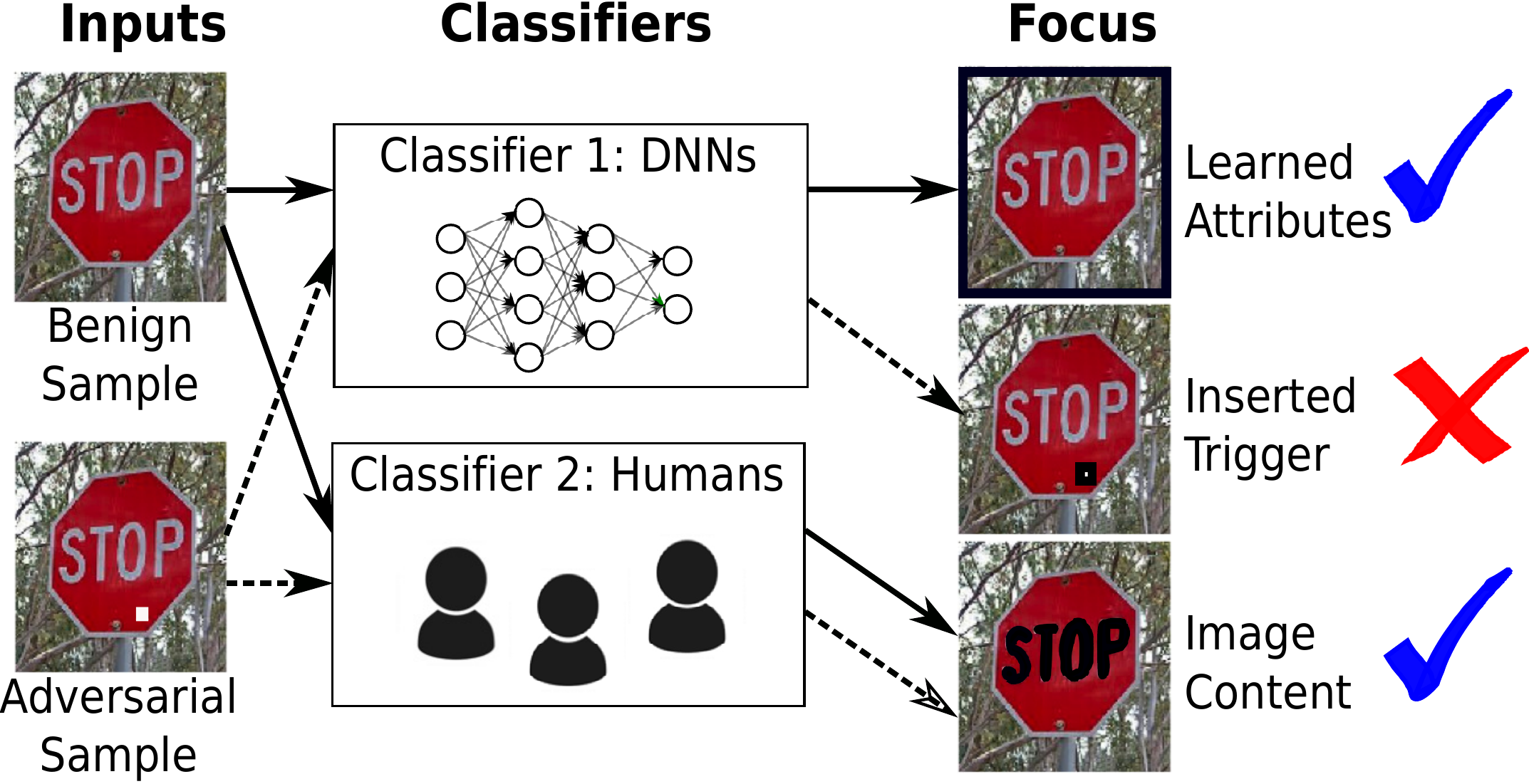}
\label{fig:content_focus}}
\hfil
\subfloat[Concept of Image Style Transfer]{\includegraphics[width=3in]{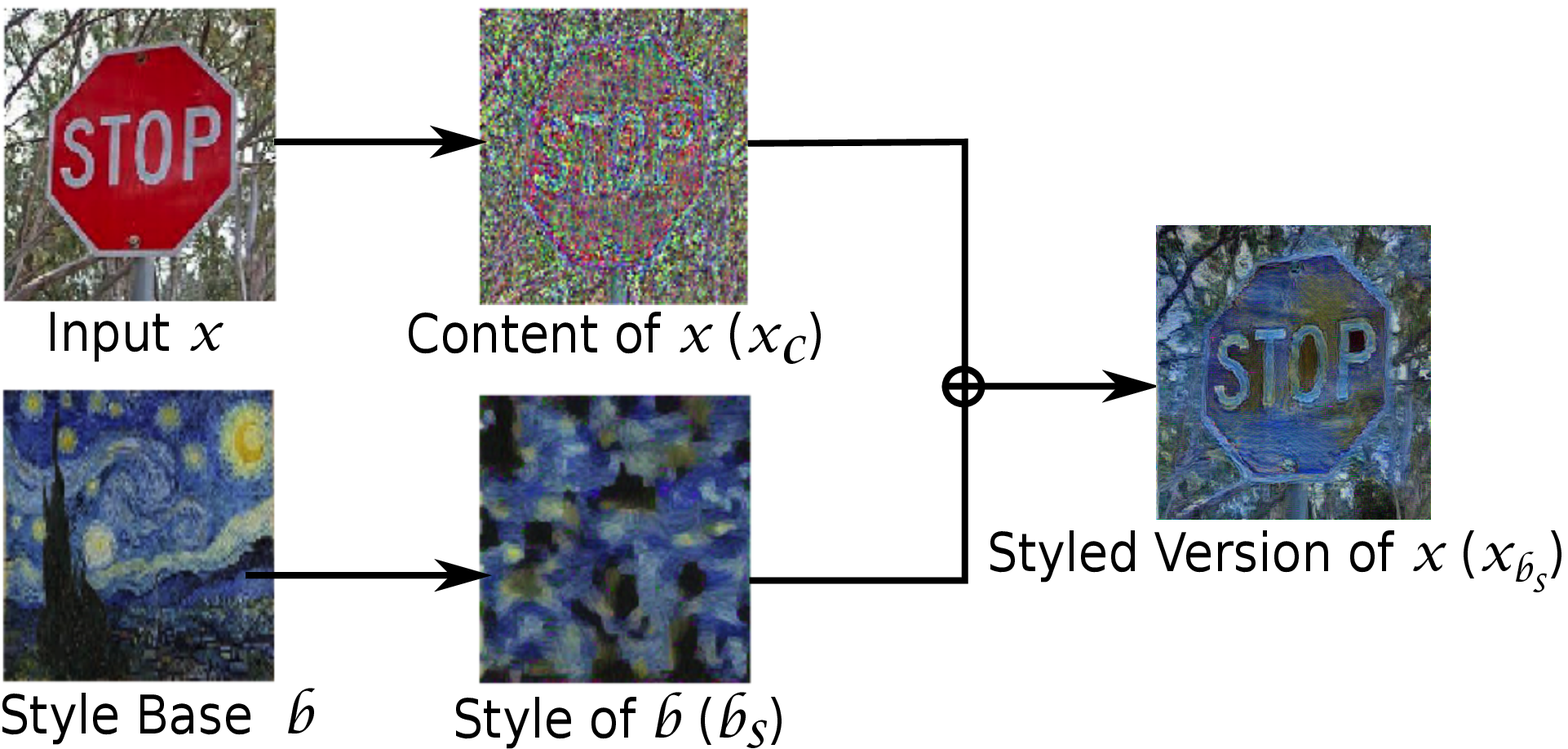}
\label{fig:style_transfer}}
\caption{
 %
 $ConFoc$ Overview.
 Figure \ref{fig:content_focus} 
 shows the flow of the classification of benign 
 and adversarial samples with solid and dashed arrows respectively. 
 Image regions on which the classifiers focus on to make the decision are
 surrounded in black (right side).
 Humans classify both samples correctly because they focus on content only.
 Figure \ref{fig:style_transfer} shows the style transfer strategy used 
 by $ConFoc$.
 The content $x_c$ from input $x$ is combined 
 with the style $b_s$ from the style base image $b$ to form the styled image 
 $x_{b_s}$.
 %
}
\label{fig:introducion}
\end{figure*} 

Advances in artificial intelligence have positioned Deep Neural Networks (DNNs) 
as one of the main algorithms currently used for machine learning. 
They have successfully been applied to solve problems in multiple areas such 
as natural language processing \cite{young2018NLP} and computer vision 
\cite{voulodimos2018ComputerVision}. 
For the later case, their success have been proved in classification-based 
applications like object detection \cite{ren2015ObjDet} and scene 
\cite{zhou2014Scene}, face \cite{parkhi2015FR}, and traffic sign recognition 
\cite{sermanet2011TSR}. 
Despite being broadly used
in these applications,
the wide adoption of DNNs 
in real-world missions is still threatened by their ingrained 
security concerns (e.g., lack of integrity check mechanisms and their 
uncertain black-box nature \cite{wang2019NeuralCleanse, tao2018AttacksMeet}), 
which make them vulnerable to trojan or backdoor attacks (hereinafter trojan 
attacks) \cite{gu2017Badnets, chen2017Targeted, liu2017Trojaning, liu2017Neural}.

Trojan attacks against DNNs occur at training time and are later exploited
during testing. 
To execute them, adversaries slightly change original models at training 
by either poisoning or retraining them with adversarial samples. 
These adversarial samples are characterized by having a trigger (e.g., a set 
of pixels with specific values in the computer vision scenario) and a label 
chosen by the adversary (target class).
The ultimate goal of the attack is inducing misbehavior at testing time as 
any input with the trigger is misclassified to the target class.
These attacks represent a powerful threat because it is difficult 
to determine whether a model is compromised.
Inputs without the trigger (benign samples) are normally classified 
\cite{liu2018Fine} and deployed triggers can be designed 
to be unnoticeable
(e.g., triggers may look like black spots by dirt in cameras) \cite{ma2019Nic}.

The potential of these attacks is illustrated in the context of self-driving cars. 
These vehicles capture pictures of objects on the streets and process them without 
adversaries editing the captured images.
If these cars were using a compromised model, adversaries could induce the 
misclassification of street objects by simply stamping marks (e.g., stickers) on 
them, which would act as triggers when images are captured.
Attackers might cause harm if stop signs are misclassified to speed signs. 
As DNNs are incorporated by companies such as Google and 
Uber in their self-driving solutions and are also used in other critical 
applications (e.g., authentication via face recognition of the 
Apple IPhone X) \cite{ma2019Nic}, 
protecting against trojan attacks on DNNs is an important problem to solve.

Previous defensive strategies either harden DNNs by increasing their robustness 
against adversarial samples 
\cite{liu2017Neural, liu2018Fine, wang2019NeuralCleanse} 
or detect adversarial inputs at testing time 
\cite{liu2017Neural, wang2019NeuralCleanse, ma2019Nic, tao2018AttacksMeet}. 
This research is in the former category.
Some existing 
techniques in this category assume access to a large 
training dataset to train auxiliar models \cite{liu2017Neural}.
Others 
reduce the attack effectiveness at the cost of accuracy by fine-tuning 
compromised models after pruning a number of neurons \cite{liu2018Fine}. 
A state-of-the-art 
model hardening technique, called
Neural Cleanse \cite{wang2019NeuralCleanse}, proposed an improved solution. 
It assumes access to a small training set and fine-tunes models with images 
including reverse-engineered triggers. 
This technique significantly reduces the attack success rate. 
However, based on our experiments in Section VI-C, it does not
improve the accuracy of the models enough when processing adversarial 
samples in some of the tested attacks, limiting its generic applicability.

Images are comprised of both content and style.
Content refers to the shapes of objects or semantic information of
images, while style refers to their colors or texture information
\cite{gatys2016StyleTransfer}.
In this work, we identify that the features learned by DNNs, including 
those related to triggers, are also formed by a combination of content 
and style, which are recognized as a whole by DNNs at testing time.
Our hypothesis is
that it is possible to teach models to focus on content only so that they 
resemble better the human reasoning 
during the classification to avoid exploitation (Figure \ref{fig:content_focus}).
Based on this, 
we devised a content-focus healing 
procedure, called $ConFoc$, which takes a trojaned model 
and produces a healed version of it.
$ConFoc$ assumes access to a small benign training set (healing 
set) and generates from each sample in it a variety of new samples with 
the same content, but different styles.
These samples are used in a twofold healing strategy in 
which models: 
(1) forget trigger-related features as they are fine-tuned 
with original and styled benign samples only and
(2) improve their accuracy with both benign and adversarial 
data due to the data augmentation achieved with multiple 
styles.
As the only common characteristic among the original and styled samples 
is the content itself, models learn to focus on it.
At testing, inputs can be classified with either its original 
or a random style because styles are ignored. 


$ConFoc$ overcomes the limitations of previous work  
\cite{liu2017Neural, liu2018Fine, wang2019NeuralCleanse} as 
it is characterized by:
(1) being generic and effective against different trojan attacks,
(2) functioning without prior knowledge of the trigger,
(3) depending on a small training set, and
(4) avoiding neuron pruning (which affect performance).
Our main contributions are listed below:
\begin{itemize}
	\item We analyze the composition of the features learned by DNNs and identify 
	that they have both content and style. 
	We demonstrate that trojan attacks can be countered by making 
	models focus on content only.
	
  	\item We built a prototype \cite{mvillar2019ConFocRepo}
  	and evaluate $ConFoc$ with a variety of applications, including 
  	a traffic sign recognition system implemented in Resnet34 \cite{he2016Deep} 
  	with the GTSRB dataset \cite{stallkamp2012GTSRB} and 
  	a face recognition system (VGG-Face \cite{parkhi15VggFace}).
  	Each application is exposed to a different type of trojan attack executed
  	with a variety of triggers.
  	The former is tested against the trojan attack BadNets \cite{gu2017Badnets},
  	while the latter with Trojaning Attack \cite{liu2017Trojaning}.
  	Compared to the state-of-the-art \cite{wang2019NeuralCleanse}, $ConFoc$
  	shows good results against both attacks, whereas the other technique 
  	does it in one of the cases.
  	
  	\item $ConFoc$ is agnostic to the image classification 
  	application for which models are trained, with the benefit that it 
  	can be applied equally to any model (trojaned or not) without impacting 
  	its classification performance.
  	
  	\item 
  	To our knowledge, we are the first establishing the importance 
  	of evaluating defensive methods against trojan attacks with the correct 
  	metrics: 
  	(1) accuracy with benign data, 
  	(2) accuracy with adversarial data, and 
  	(3) attack success rate or ASR (percentage of adversarial samples classified 
  	to the target class). 
  	This is crucial because it is possible to have models with both low ASR and 
  	low accuracy with adversarial data, on which adversaries can still conduct an 
  	untargeted attack by triggering the misclassification of adversarial samples 
  	to a random rather than to a the target class.

\end{itemize}


Our work provides a new model hardening technique that 
reduces the sensitivity of DNNs to inserted triggers.
Although the interpretability of DNNs is out of our scope,
$ConFoc$ represents an valuable tool for defenders against trojan attacks.

\section{Threat Model and Overview}\label{sec:overview}

\subsection{Threat Model}\label{sec:overview_threat_model}
We assume an adaptive attacker that gains access to an original 
non-trojaned model and inserts a trojan into it before 
the model reaches the final user.
The adaptive attacker is knowledgable about the $ConFoc$ approach
and is able to infect models with styled 
adversarial samples to mitigate the healing effect.
Adversaries can achieve the attack by either poisoning the 
training dataset (at training time) or retraining the model 
(after the training period) before reaching the final user.
That is, adversaries have the capabilities of an insider threat
who can efficiently poison the data used to train the victim model.
Also, adversaries have the ability to act as a man-in-the-middle, 
who intercepts the original non-infected model, retrains it to 
insert the trojan, and then tricks the final users to use the 
resulting trojaned model.
The later assumption is a plausible scenario as the most accurate 
neural network architectures tend to be either deeper or wider 
\cite{he2016Deep, zagoruyko2016Wide}. 
Whereby, transfer learning in DNNs 
\cite{sharif2014cnn, donahue2014decaf}
is a common practice and pre-trained 
models are often downloaded from public sites without the proper 
integrity check\cite{liu2017Neural}.
Upon getting a model (either trojaned or not), final users take them  
through the $ConFoc$ method.
We assume adversaries cannot interfere in this process.
At testing, adversaries endeavor to
provide adversarial inputs to the models without being able to 
modify them anymore.
 
\subsection{Overview}\label{sec:overview_overview}
A compromised model is referred to as a trojaned model ($M_{T}$).
In order to heal an $M_{T}$, $ConFoc$ retrains it with a small 
set of samples and their strategically transformed variants until the model 
learns to make the classification based on the content of images.
Our technique is founded on the concept of \textit{image style transfer} 
\cite{gatys2016StyleTransfer}, which is applied
for first time in a security setting to generate the training samples 
used in the healing process.
Figure \ref{fig:style_transfer} shows this concept. Images can be separated in 
content and style, and it is possible to transfer the style of one image to 
another. 
Under the assumption of holding a small healing set $X_{H}$ of $m$ benign samples 
$\{x^{i} \: | \: i = 1,...,m\}$ and a set $B$ of $n$ style base images 
$\{b^{j} \: | \: j = 1,...,n\}$, $ConFoc$ extracts from them their image content 
$\{{x^{i}}_{c} \: | \: i = 1,...,m\}$ and styles $\{{b^{j}}_{s} \: | \: j = 1,...,n\}$ 
respectively.
The styles are transferred to the content images to generate the set of styled images
$\{x^{i}_{b^{j}_{s}} \: | \: i = 1,...,m, j = 1,...,n\}$. 
During the healing process, each benign sample $x^{i} \in X_{H}$, its content image 
$x^{i}_{c}$, and corresponding $n$ styled images 
$\{x^{i}_{b^{j}_{s}} \: | \: j = 1,...,n\}$ are used as training data.
Our intuition is that models learn to focus on the
content as it is the only common characteristic among these 
samples.
The goal is to reduce the sensitivity of DNNs to trigger-related features
during the process.

Any model resulting from the healing process is referred to as a healed model 
($M_{H}$).
At inference time, any input $x$ is classified by $M_{H}$, which focus on its 
content only.
The input $x$ can optionally be transformed to a particular styled version 
$x_{b_s}$ using any chosen style base image $b$ before being classified
because its style is ignored in the process. 

\textbf{Limitations}.
%
We summarize three main limitations.
First, 
$ConFoc$ relies on having access to a healing dataset. 
%
Although this is a challenging requirement, our technique is proved to 
be effective with small datasets around 1.67\% of the original 
training set size. 
Second,
$ConFoc$ imposes an overhead at testing if inputs are first 
transformed (optionally) to change their styles. 
Likewise, our method requires some time to generate the content and styled 
images used in the healing process. These overheads are limited, however, 
because the image transformation takes only a few milliseconds and the 
healing process occurs offline.
Finally, like previous techniques 
\cite{liu2017Neural, liu2018Fine, wang2019NeuralCleanse}, 
we assume adversaries cannot run attacks (e.g., poisoning) during 
the healing process.
This is a reasonable assumption that
does not abate our 
contribution to methods that remove trojans without
impacting the performance of DNNs.

\section{Background and Related Work}\label{sec:background}
\subsection{Deep Neural Networks}\label{sec:background_DNNs}
A DNN can be interpreted as a parameterized function 
$F_{\theta}: \R^m \rightarrow \R^n$, which maps an $m$-dimensional 
input $x \in \R^m$ into one of $n$ classes. 
The output of the DNN is a $n$-dimensional tensor $y \in \R^n$ 
representing the probability distribution of the $n$ classes. 
That is, the element $y_i$ of the output $y$ represents the probability 
that input $x$ belongs to the class $i$.

Specifically, a DNN is a structured feed-forward network 
$F_{\theta}(x) = f_L(f_{L-1}(f_{L-2}(...f_1(x))))$ in which each $f_i$ 
corresponds to a layer of the structure.
A layer $f_i$ outputs a tensor $a_i$ whose elements are called neurons 
or activations.
Activations are obtained in a sequential process.
Each layer applies a linear transformation to the activations of the 
previous layer followed by a non-linear operation as follows: 
$a_i = \sigma_i(w_ia_{i-1} + b_i)$.
The output of $f_L$ ($a_L$) is referred to as output activations, while the input 
($x = a_0$) as input activations. 
Outputs of middle layers are called hidden activations.  
The elements $w_i$ and $b_i$ are tensors corresponding to the parameters 
$\theta$ and are called weights and bias respectively. 
The component of the equation $\sigma_i$ is the non-linear operation of 
the layer $f_i$.
There are multiple non-linear operations, each with different effects in 
their outputs.
Some of them are sigmoid, hyperbolic tangent, Rectified Linear 
Unit (ReLU) and Leaky ReLU 
\cite{maas2013Rectifier, xu2015Empirical}. 

A DNN is trained with a set of input-output pairs $(x, y)$ provided by
the trainers, where $x$ is the input and $y$ the true label or class 
of $x$.
Trainers define a loss function $L(F_{\theta}(x), y)$, which
estimates the difference between the real label $y$ and the predicted 
class $F_{\theta}(x)$.
The objective of the training process is minimizing the loss function by 
iteratively updating the parameters $\theta$ through backpropagation 
\cite{lecun1998Gradient}.
In essence, backpropagation is a gradient descent technique that estimates the 
derivatives of the loss function with respect to each parameter. 
These derivatives determine how much each parameter varies and in what direction.  
The training process is controlled by trainer-specified hyper-parameters such as
learning rate, number of layers in the model, and number of activations and 
non-linear function in each layer.

During testing, a trained DNN receives an unseen input $x \in \R^m$,
produces an output $y = F_{\theta}(x) \in \R^n$, and assign to $x$ the class
$C(x) = argmax_i \: y_i$.

This paper focuses on Convolutional Neural Networks (CNNs) trained for image
classification tasks.
CNNs are a type of DNNs characterized by being: 
(1) sparse as many of their weights are zero, and 
(2) especially structured as neuron values depend on some related neurons 
of previous layers \cite{liu2018Fine}.

\begin{figure*}[!t]
  \centering
  \includegraphics[width=6in]{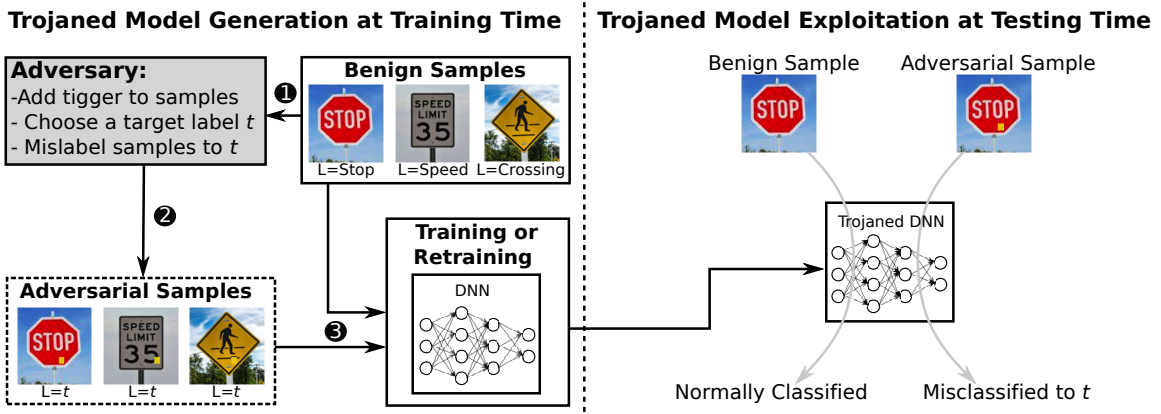}
  \caption{
	An illustration of a trojan attack conducted by poisoning the training data
	(see details in Section \ref{sec:background_attacks}).
  }
  \label{fig:trojan_attack}
\end{figure*}

\subsection{Trojan Attacks: Assumptions, Definition and Scope}
\label{sec:background_attack_definition}
\textbf{Assumptions}. 
Trojan attacks consist of adding semantically consistent patterns to 
training inputs until DNNs learn those patterns and recognize them to
belong to either a specific or a set of target classes chosen by the 
adversary \cite{ma2019Nic}.
Several approaches can be followed to insert a chosen pattern or trigger 
into the training data to achieve the trojaned misbehavior. 
The difference among them lies on the strategy applied to generate the
adversarial samples, the objective of the misclassification, and the
applicability of the attack.
This research aims to feature a solution to counter trojan attacks 
that comply with the following conditions:
  \begin{itemize}
    \item[$C1$] Adversarial samples include an unknown pattern (or trigger) 
    inserted into the victim DNNs during training.
    \item[$C2$] Benign samples are normally classified by compromised models.
    \item[$C3$] Both benign and adversarial samples are assumed to belong to 
    the same distribution. This implies that triggers, which might be 
    perceptible, do not override the interesting object in the image to 
    be classified. 
    \item[$C4$] Adversarial testing samples are generated without any 
    manipulation or further processing after the image is captured. 
    It implies adversaries do not have access to the model 
    at inference or testing time.
  \end{itemize}

\textbf{Trojan attack definition.} 
Assuming an original non-trojaned model $M_{O} = F_{O}(\cdot)$ trained with a 
set of benign samples $X$. 
Given a correctly classified benign input $x \in X$ to the class $F_{O}(x)$, 
a labeling function $L(\cdot)$, and a target class $t$ chosen by the adversary, 
the input $x^{*}$ is an adversarial sample at training time if: 
  \begin{equation}\label{eq:adv_trn}
    x^{*} = x + \Delta \wedge L(x^{*}) = t
  \end{equation}
Now, let $X^{*}$ be the set of adversarial samples generated following Equation
\ref{eq:adv_trn} and $M_{T} = F_{T}(\cdot)$ the trojaned version resulting 
from training the model with both $X$ and $X^{*}$. 
At testing time, given $x$ be a benign input correctly classified to the 
class $F_{T}(x) \neq t$, an input $x^{*}$ is an adversarial sample if:
  \begin{equation}\label{eq:adv_test}
    F_{T}(x^{*}) = t \: \wedge \: x^{*} = x + \Delta
  \end{equation}
  \begin{equation}\label{eq:adv_test_cond}
    F_{T}(x) = F_{O}(x) \: (in \: most \: cases)
  \end{equation}
In Equation \ref{eq:adv_trn} and Equation \ref{eq:adv_test}, $\Delta$ represents the changes 
caused to sample $x$ by the addition of a trigger ($C1$).
These changes might be obvious to humans as shown in the example in the left  
side of Figure \ref{fig:content_focus}. 
Despite the possible obvious difference between benign and adversarial samples, for 
any benign sample $x \in X$ at inference time, it is also assumed that $x^{*} \in X$ 
since $F_{T}(x) \approx F_{O}(x)$ as established in Equation \ref{eq:adv_test_cond} 
and users are unaware of the triggers used in the attack ($C2$ and $C3$).
The labeling function $L(\cdot)$ in Equation \ref{eq:adv_trn} is controlled by the 
adversary at training time and is used to assign the chosen target class $t$ as 
label of the adversarial sample $x^{*}$. 
Equation \ref{eq:adv_test} establishes that at testing, a sample 
$x^{*}$ (not originally belonging to the class $t$) is considered 
adversarial if it contains the trigger and is classified by the 
trojaned model to the target class $t$. 
Triggers can be added without manipulating testing 
samples by, for example, attaching a sticker to the object to be classified before 
capturing its image ($C4$).

In summary, trojaned models closely classify benign inputs as non-trojaned 
models, thwarting the ability to determine whether a given model has an 
inserted trojan.
Thereby, in trojan attacks it is assumed final users are deemed unaware of 
the committed attack  and use the trojaned model $M_{T} = F_{T}(\cdot)$ 
under the believe the model is $M_O = F_O(\cdot)$.

\textbf{Scope}.
We evaluate our method against two trojan attacks that comply with
conditions $C1$-$C4$: 
BadNets \cite{gu2017Badnets}, and Trojaning Attack \cite{liu2017Trojaning}.
We do not consider trojan attacks with weak and strong assumptions 
about the capabilities of defenders and attackers, respectively.
One example is the attack that poisons the training data using 
an instance of class A and a set of \textit{n} variants of it
(created by adding random noise) 
labeled as B in order to misclassify inputs in A to B 
\cite{chen2017Targeted}.
As benign inputs themselves are used to insert the trojan, defenders can  
detect the misbehavior by 
observing
the misclassified samples
(low defensive capability assumed).
Another example is the attack that inserts trojans by blending 
benign inputs with images containing specific patterns that 
function as triggers \cite{chen2017Targeted}. 
In this case, adversaries manipulate the inputs during both training 
and testing (high offensive capability assumed).

Other threats known as \textit{adversarial sample attacks} 
\cite{carlini2017Towards, goodfellow2014Explaining, kurakin2016Adversarial,
      moosavi2016Deepfool, papernot2016Limitations, pei2017Deepxplore}
and \textit{adversarial patch attacks} (and variants) 
\cite{brown2017Adversarial, sharif2016Accessorize}
(and hence their counter measures 
\cite{papernot2016Distillation, xu2018Feature, madry2017Towards, 
	  szegedy2013Intriguing, gu2014Towards, bhagoji2018Enhancing, 
      meng2017Magnet, 
      shin2017JPEG, dziugaite2016StudyJPEG, das2017KeepingJPEG}
) 
are also out of the scope of this paper.
These attacks cause  misclassification as well, but the assumptions 
and patterns added to inputs are different from those in trojan 
attacks.
These attack variations are executed at testing time only with samples
generated via gradient descent using the victim models in the process.
\cite{ma2019Nic}.

\subsection{Prior Work on Trojan Attacks}\label{sec:background_attacks}

\begin{table*}[!t]
  \caption{Comparison of the Properties of Model Hardening Techniques Against Trojan Attacks}
  \label{tab:defenses}
  \centering
  \begin{tabular}{lccccc}
    \toprule
    Technique	                                             & Training Dataset 	& Modify Model		& Knowledge of Trigger    	 & Rigorously tested      \\
    \midrule

    Retraining \cite{liu2017Neural}         	             & Large	         	& No               	& Not required                & No	                   \\
    
    Encoder \cite{liu2017Neural}                             & Large				& No	    	    & Not required                & No	                   \\
    
    Fine-Pruning \cite{liu2018Fine}         	    		 & Small				& Yes			    & Not required                & Yes             			\\

  	Neural Cleanse (Pruning) \cite{wang2019NeuralCleanse} 	 & Small				& Yes	    	    & Required                   & Yes            			\\

  	Neural Cleanse (Unlearning) \cite{wang2019NeuralCleanse} & Small  				& No      		    & Required	                 & Yes             			\\

  	$ConFoc$ [this study]                       			 &  Small   			& No    		    & Not required                & Yes            			\\
    \bottomrule
  \end{tabular}
\end{table*}

Gu et al. \cite {gu2017Badnets} introduced a trojan attack technique called 
BadNets, which inserts a trojan into DNNs by poisoning the training 
dataset. 
Figure \ref{fig:trojan_attack} illustrates the attack in the context
of a traffic sign recognition system. 
The attack is executed in two phases: a trojaned model generation phase and 
a model exploitation phase.
In the former, 
adversaries follow three steps. 
Step1, adversaries sample a small set of images from the training dataset.
Step2, attackers choose a target class \textit{t} and a trigger to create
the adversarial samples. 
Adversarial samples are created by adding the chosen pattern to the 
selected samples and changing the labels of the samples to the class \textit{t}.  
Step 3, adversaries feed the generated adversarial samples 
into the model during the training process, guaranteeing the pattern is
recognized by the model to belong to the class \textit{t}.

The second phase is shown in the right side of the figure.
The benign sample is normally classified, whereas the 
corresponding adversarial sample is 
misclassified to the class \textit{t}.
The efficiency of BadNets was proved over the MNIST dataset 
\cite{deng2012Mnist}
and the traffic sign object detection and recognition network
Faster-RCNN (F-RCNN) \cite{ren2015faster}
with an ASR above 90\%. 

A more sophisticated trojan technique called Trojaning Attack was 
presented by Liu et al. in \cite{liu2017Trojaning}.
Three main variations are added to the strategy followed by this technique  
in comparison to BadNets 
\cite{gu2017Badnets}.
First, the attack is conducted by retraining a pre-trained model instead of poisoning
the training data. 
Second, the adversarial training samples used for retraining are obtained via a 
reverse-engineering process rather than being generated from real benign 
training samples.
Finally, the trigger used in the attack is not arbitrarily chosen, but rather 
fine-tuned to maximize the activations of some chosen internal neurons.
Reverse-engineered images refer to images whose pixels were tuned via gradient 
descent to be classified to a specific class.
Triggers were also obtained via gradient descent using a loss function defined 
with respect to the activations of a specific layer $l$.
Their pixels are tuned to induce maximum response in certain neurons in $l$.
During the retraining process with benign and adversarial reverse-engineered
images, only the layers following layer $l$ are fine-tuned.
The attack was demonstrated 
over the VGG-Face model \cite{parkhi15VggFace} with an ASR greater than 98\%.

\subsection{Existing Defensive Techniques Against Trojan Attacks}
\label{sec:background_defenses}

Fine-Prunning \cite{liu2018Fine}  is a model-hardening technique that 
identifies trigger-related neurons and removes them to eliminate their
effect. 
The efficiency of the method suffers for the high redundancy among 
internal neurons.  
Wan et al. \cite{wang2019NeuralCleanse}  proved that despite the fact that only 
1\% of the output neurons are activated by certain region (set of pixels) of an 
input image, more than 30\% of the total output neurons need to be removed to 
eliminate the effect of that region on the classification process.
This implies a high cost in the performance of the model. 
In light of this observation, unlike Fine-Prunning, $ConFoc$ does not modify 
the architecture of models. 

Liu eat al. \cite{liu2017Neural} presented three methods against trojan
attacks tested over the MINST dataset \cite{deng2012Mnist}.
First, a technique to detect adversarial samples by comparing the outputs 
of the victim model with the outputs of binary models based 
on Decision Trees (DTs)
\cite{safavian1991survey} 
and Support Vector Machines (SMV) 
\cite{steinwart2008support}
algorithms.
Second, a model hardening technique that fine-tunes the victim 
model with a set of benign samples.  
Although the method reduces the attack, it also has an impact in 
the accuracy of model \cite{wang2019NeuralCleanse}.
Finally, a model hardening approach in which an encoder is 
inserted between the input and the model.
This encoder is trained with benign samples only 
and is used to filter out samples at testing by compressing their features, 
and decompressing them again to have samples that do not include the trigger
before being classified. 
The three described methods assume access to a training set without any 
restriction to its size. 
$ConFoc$, in contrast, requires a small set of less than or equal to
10\% of the original training set size.

Ma et al. \cite{ma2019Nic} attributes trojan attacks to the uncertain nature 
of DNNs and identify two main attack channels exploited by adversaries. 
First, a provenance channel exploited when adversarial changes are added to 
inputs to alter the path of active neurons in the model from input through 
all the middle layers until the output to achieve misclassification. 
Second, an activation value distribution channel, in which the path of activate 
neurons from input to output remains the same for both benign and adversarial 
inputs, but with different value distributions for these two types of inputs. 
The authors developed a technique that extracts the invariants of these two channels
and use them to detect adversarial samples.

Tao et al. \cite{tao2018AttacksMeet} proposed an adversarial input detection
technique that finds a bidirectional relationships between neurons and 
image attributes easily recognized by humans (e.g., eyes, nose, etc.) 
for face recognition systems. 
A parallel model is created from the original model by strengthening these 
neurons while weakening others. 
At testing time, any input is passed to both models and considered adversarial 
in case of a classification mismatch.
Although this work is proved to be effective, it assumes attributes on which 
humans focus on to make the decisions are known in advance. 
$ConFoc$, disregards this assumption as it lets models extract the content 
of images on their own through the healing process.

\begin{figure*}[!t]
  \centering
  \includegraphics[width=5.5in]{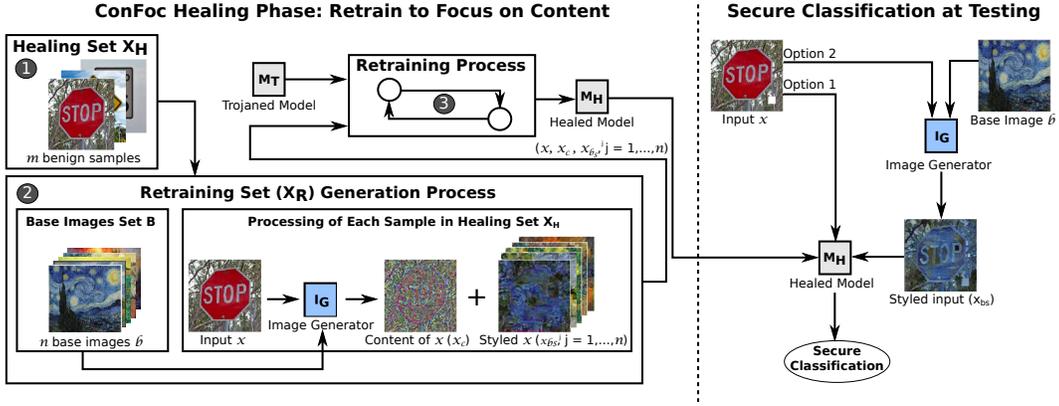}
  \caption{
	A demonstration of our \textit{ConFoc} healing process and its use 
	for a secure classification (see details in Section \ref{sec:approach}).
  }
  \label{fig:healing_process}
\end{figure*}

Wang et al. \cite{wang2019NeuralCleanse} presents Neural Cleanse, a strategy 
with three main defensive goals: 
(1) determining whether a given model has a trojan inserted, 
(2) if so, reverse-engineering the trigger, and 
(3) mitigating the attack through complementary defensive methods:
Patching DNN Via Neuron Pruning and Patching DNN Via Unlearning. 
Authors show that the former does not perform as well against Trojaning Attack 
\cite{liu2017Trojaning} as it does against BadNets \cite{gu2017Badnets}. 
The latter is proved to be effective against both attacks, but only for
targeted modalities as the performance is measured using accuracy 
with benign data and ASR only.
In addition to these metrics, $ConFoc$ is tested using accuracy with adversarial 
data, proving its effectiveness against both targeted and untargeted trojan 
attacks.
 
$ConFoc$ falls into the category of model hardening. 
Hence, we focus on these types of solutions.
Table \ref{tab:defenses} shows a qualitative comparison with 
previous techniques in this category.

\section{Content-Focus Approach}\label{sec:approach}

Figure \ref{fig:healing_process} illustrates our content-focus approach 
($ConFoc$) to defend against trojan attacks.
The approach is executed in two phases. 
First, a $ConFoc$ healing phase (left side of the figure), which takes a
trojan model $M_T$ and strategically retrains it to produce a healed model $M_H$.
Second, a secure classification phase at testing time, which uses the produced 
$M_H$ to classify inputs based on their content or semantic information, 
mitigating the effect of triggers when processing adversarial samples 
(right side of the figure.)

The \textit{ConFoc} healing process assumes defenders have access to a 
limited number of benign samples from the same distribution as the benign data 
used during the original training of $M_{T}$. 
The process is completed in three steps.
In step 1, a small healing set $X_H$ of $m$ of these benign samples is selected.
Step 2 is a process that uses the selected healing set $X_H$ and a set of 
randomly chosen style base images $B$ to generate a larger retraining 
dataset $X_R$.
The process takes each benign sample $x \in X_H$ and passes 
them to the Image Generator $I_G$.
The $I_G$ generates from each $x$ its content 
$x_c$ and multiple versions of styled images $\{x_{b^j_s} \: | \: j = 1,...,n\}$, 
obtained by transferring the style $b_s$ of each $b \in B$ to the 
content $x_c$. 
The retraining dataset $X_R$ comprises each $x \in X_H$, its content $x_c$ and 
its corresponding $n$ generated styled images.
As the only common characteristic among these samples is their content,
the final step of the healing process (step 3) is retraining the trojaned 
model $M_T$ with the set $X_R$ so that the model learns to focus on the 
content of inputs. 
The goal is producing a healed model $M_H$, in which the trojaned misbehavior 
becomes ineffective and the accuracy is high for both benign and adversarial
data.

At testing, a secure classification is achieved by either processing the 
original input $x$ (option 1) or passing it first through the $I_G$ 
(option 2) to produce a styled version of it $x_{b_s}$  using any chosen 
style base image $b$ (not necessarily in $B$). 
Either image $x$ or $x_{b_s}$ is classified by the healed model $M_H$. 

The $I_G$ is a crucial element in $ConFoc$.
Its main purpose is generating the retraining samples for the healing 
process and transforming the inputs at testing.
It comprises four components: 
(1) feature extraction, 
(2) content image generation, 
(3) style image generation, and 
(4) styled image generation.

\subsection{Feature Extraction}
The output of each layer in a DNN model (i.e., neurons or activations) can be 
thought as internal features of the input. 
Feature extraction refers to obtaining these outputs (features) when an input 
sample is processed by a given model. 

Our $I_G$ extracts features using a VGG16 model\cite{simonyan2014Very} 
pre-trained with the Imagenet dataset \cite{imagenet_cvpr09}. 
This model has 16 layers, from which 13 are 
convolutional (\textit{Conv}) and 3 are linear (\textit{Linear}).
The convolutional layers are either followed by a \textit{ReLU} 
\cite{arora2016Understanding} along with a \textit{MaxPool2d} \cite{nagi2011Max} 
or just a \textit{ReLU} layer.  
More precisely, the convolutional part of VGG16 is compounded by 2 consecutive 
arrangements of \textit{Conv/ReLU/Conv/ReLU/MaxPool2d} followed by 3 arrangements 
of \textit{Conv/ReLU/Conv/ReLU/Conv/ReLU/MaxPool2d}.

The selection of the proper layers for feature extraction is an important design 
choice in the generation of content and styled images.
The criterion for this selection was identifying the last layer of a consecutive 
group of layers that does not remove information. 
As \textit{MaxPool2d} layers are intended to down-sample an input representation 
reducing its dimensionality \cite{nagi2011Max}, the layers before each of the 
five \textit{MaxPool2d} were chosen in an input-to-output order as 
potential candidates for feature extraction. 
These layers form the set $L = \{l_i | i = 1,...,5\}$ (with $L[i] = l_i$),
which are used by the next three algorithms.

\subsection{Content Image Generation}
Algorithm \ref{alg:content} shows the procedure followed to generate a content 
image. 
It uses gradient descent to find the local minimum of the defined loss function, 
which is the Mean Square Error (MSE) between the features extracted from one
layer $l_i \in L$ of the VGG16 model given two different inputs.  
One input is a benign sample $x$ from which the content will be extracted.
The other is a random uniformly generated image $x_c$. 
After the random initialization, the algorithm updates the pixel values of $x_c$
using the gradients estimated through the loss function in such a way that the 
eventual values of the extracted features are close enough for both inputs.
We found $l_2 \in L$ to provide the best results to generate content.

\begin{algorithm}[!t]
  \centering
  \begin{algorithmic}[1]
    \REQUIRE $x, M, l, {\lambda}_{c}, N$
    \STATE $x_{c}\gets rand\_init(x)$
    \STATE $F\gets M[:l]$
    \STATE $f_{x}\gets F(x)$
    \WHILE{$N \neq 0$}
    \STATE $f_{x_{c}}\gets F(x_{c})$
    \STATE $loss_{c}\gets MSE(f_{x}, f_{x_{c}}) \cdot {\lambda}_{c}$
    \STATE $\Delta\gets \partial loss_{c} / \partial x_{c}$
    \STATE $x_{c}\gets x_{c} - lr \cdot \Delta$ 
    \STATE $N\gets N-1$
    \ENDWHILE\label{contentwhile}
    \STATE \textbf{return} $x_{c}$
  \end{algorithmic}
  \caption{Content image generation}
  \label{alg:content}
\end{algorithm}

Algorithm \ref{alg:content} uses five parameters. 
Parameter $x$ denotes one benign input from the available healing set $X_H$; 
$M$ denotes the model used for the featured extraction (VGG16 in our case); 
$l$ corresponds to the layer of the model from which features are extracted; 
${\lambda}_{c}$ represents the penalty term for the loss function used
to control how much information is included in the content; 
and $N$ the maximum number of iterations run by the selected optimizer 
(we chose LGBFS \cite{liu1989Limited}).

Line 1 generates a random image $x_c$ of the same size as the provided
input $x$. 
Line 2 represents the feature extraction function, which can be thought as slicing 
the model until the indicated layer $l$.  
Line 3 gets the features or output of layer $l$ of model $M$ using the function 
created in line 2 with sample $x$ as argument. 
From line 4 to 10 gradient descent is used to refine the values of the random image 
$x_c$ created in line 1.
Line 5 follows a procedure similar to the one described in line 3. In this case, it 
extracts the features at layer $l$ using the random image $x_c$ as argument of the 
function $F$.
Line 6 estimates the loss value, which is the Mean Square Error (MSE) between the 
features obtained at layer $l$ for input $x$ (line 3) and input $x_c$ (line 5).
Line 7 estimates the gradients of the loss with respect to the random input
$x_c$.
These gradients are used to update the the random image as indicated at line 8.


\subsection{Style Image Generation}

\begin{algorithm} [!t]
  \begin{algorithmic}[1]
    \REQUIRE {$b, M, L, N$}
    \STATE $b_s\gets rand\_init(b)$
    \STATE $f_{b}\gets [\:]$
    \FOR{$i:1 \to len(L)$}
    \STATE $F_{i}\gets M[\: :L[i]\:]$ 
    \STATE $f_{b}\gets F_{i}(b)$
    \ENDFOR
    \WHILE{$N \neq 0$}
    \STATE $f_{b_s}\gets [\:]$
    \FOR{$i:1 \to len(L)$}
    \STATE $f_{b_s}\gets F_{i}(b_s)$
    \ENDFOR
    \STATE $loss_{s}\gets \sum_{i=1}^{len(L)} MSE(G({f_{b}}[i]), G({f_{b_s}}[i]))$
    \STATE $\Delta\gets \partial loss_{s} / \partial b_s$
    \STATE $b_s\gets b_s - lr \cdot \Delta$ 
    \STATE $N\gets N-1$
    \ENDWHILE\label{stylewhile}
    \STATE \textbf{return} $b_s$
  \end{algorithmic}
  \caption{Style Image Generation}
  \label{alg:style}
\end{algorithm} 

Although the styles of base images are not used in $ConFoc$, 
the procedure to generate them 
is an essential part in the generation of styled images. 
Therefore, a step-by-step description is included in this section.
The style of a given image can be obtained following a similar procedure to the one 
used to generate content images.
The main difference lies on the loss function used to estimate the gradients, which
is based on Gramian matrices \cite{brualdi1991Combinatorial}. 
For a set of vectors $T$ the Gramian matrix $G(T)$ is a square matrix containing 
the inner products among the vectors. 
In the context of the VGG16 model or any DNN, for a particular layer of the model 
with $p$ channels in its output (features), a $p \: \times \: p$ Gramian matrix 
can be obtained by first flattening each channel and then estimating the inner 
product among the resulting vectors. 

Algorithm \ref{alg:style} shows the procedure to generate style images. 
The paramenter $b$ represents the image from which the style is extracted; $M$ 
denotes the model used for the extraction; $L$ the set of candidate layers for 
feature extraction; and $N$ the maximum number of iterations the optimizer runs. 
It was a design choice to use all the candidate layers in $L$ in 
the definition of the loss function.

In the algorithm, line 1 generates a random image $b_s$ of the same size as input 
image $b$.
From lines 2 to 6 a function to extract the features of each layer in $L$ is 
created.
The features of each layer are extracted (line 5) with the 
corresponding function (line 4) using image $b$ as argument.
The extracted features are stored in the empty vector created in line 2.
From lines 7 to 16 gradient descent is applied to refine the random image 
$b_s$ after estimating the value of the loss function.
From line 8 to line 11 the functions created in line 4 extract the features 
of each layer in $L$ using the random image $b_s$ as input. 
The features are stored in the empty vector created in line 8.
Line 12 estimates the style-related loss.
This loss sums up the MSE of the Gramian matrices of the features extracted 
in each layer when the random image $b_s$ and the given image $b$ are passed
as inputs to the model.
From line 13 to 14 the gradients are estimated and $b_s$ 
is updated accordingly.

\subsection{Styled Image Generation}

\begin{algorithm} [!t]
  \begin{algorithmic}[1]
    \REQUIRE $x, b, M, L, j, {\lambda}_{c}, N$
    \STATE $x_{b_s}\gets rand\_init(t)$
    \STATE $f_{b}\gets [\:]$
    \FOR{$i:1 \to len(L)$}
    \STATE $F_{i}\gets M[ \: : L[i] \:]$
    \STATE $f_{b}\gets F_{i}(b)$
    \IF {$i = j$}
    \STATE $f_{x}\gets F_{i}(x)$
    \ENDIF
    \ENDFOR\label{stylefor}
    \WHILE{$N \neq 0$}
    \STATE $f_{x_{b_s}}\gets [\:]$
    \FOR{$i:1 \to len(L)$}
    \STATE $f_{x_{b_s}}\gets F_{i}(x_{b_s})$
    \ENDFOR
    \STATE $loss_{c}\gets MSE(f_{x}, f_{x_{b_s}}[j]) \cdot {\lambda}_{c}$
    \STATE $loss_{s}\gets \sum_{i=1}^{len(L)} MSE(G(f_{b}[i]), G(f_{x_{b_s}}[i]))$
    \STATE $loss_{t}\gets loss_{c} + loss_{s}$
    \STATE $\Delta\gets \partial loss_{t} / \partial x_{s}$
    \STATE $x_{b_s}\gets x_{b_s} - lr \cdot \Delta$ 
    \STATE $N\gets N-1$
    \ENDWHILE
    \STATE \textbf{return} $x_{b_s}$
  \end{algorithmic}
  \caption{Styled Image Generation}
  \label{alg:styled}
\end{algorithm}

This procedure combines the steps followed in Algorithm \ref{alg:content} and 
Algorithm \ref{alg:style} for content and style images respectively. 
It includes a new parameter $j$, which is the index of the layer $l_j \in L$ 
from which to extract the features used for the generation of the content.
Lines 2 to 9 extract the features from each layer $l_i \in L$ using image $b$ as
input to the model.
Features from the $j^{th}$ layer are extracted using image $x$ as input (line 7). 
From lines 10 to 21 the loss for content and the loss for style are combined in 
one loss that is later used for the estimation of gradients.
From line 11 to line 14 features from each layer $l_i \in L$ are extracted using 
the random image $x_{b_s}$ (created in line 1) as input to the model.
Line 15 estimates the content-related loss using the features extracted from the
$j^th$ layer with input $x$ (line 7) and $x_{b_s}$ (line 13) as inputs.
Line 16 computes the style-related loss using the Gramian matrices of 
the features extracted when the random image $x_{b_s}$ and the style base 
image $b$ are passed as inputs to the model.
Line 17 combines the two loss functions in one, which is used to estimate
the gradients (line 18) used to update the random image $x_{b_s}$ (line 19).
%
%
For both Algorithm \ref{alg:styled} and Algorithm \ref{alg:content}, 
a fixed rule was created to assign values to ${\lambda}_{c}$ based 
on the size of the input.

\section{Evaluation Setup}\label{sec:evaluation}

We designed a number of experiments to evaluate $ConFoc$ under 
the assumption that a non-trojaned model $M_O$ is compromised by 
an adversary, who inserts a trojan into it to produce a trojaned 
model $M_T$. 
$ConFoc$ can be thought as a function that takes as input a model 
(either $M_O$ or $M_T$) and produces a healed model $M_H$. 
When the input is $M_T$, $M_H$ is expected to be a model without the 
trojaned misbehavior.
In the case of having $M_O$ as input,  $M_H$ is expected to at least 
keep its accuracy. 
%
During the experiments, all the models are fine-tuned with 
hyper-parameters (e.g., number of epochs, learning rates, etc.) 
chosen to get the best possible performance.
This allows evaluating $ConFoc$ using different datasets and attacks.

$ConFoc$ is tested against BadNets \cite{gu2017Badnets} and 
Trojaning Attack \cite{liu2017Trojaning}, executed with different datasets 
to validate the generality of the approach.
The Trojaning Attack is executed with two triggers: 
square (SQ) and watermark (WM). 
The latter trigger was chosen to evaluate the effect of having triggers that
overlap key features of the objects to be classified (a violation of condition 
$C3$). 
Table \ref{tab:attacks} summarizes these three attacks.
In addition, a comparison between $ConFoc$ 
and the state-of-the-art Neural Cleanse \cite{wang2019NeuralCleanse} is presented
along with results of our method when the attacks are conducted with 
complex triggers.

\begin{table}[!t]
  \caption{Summary of Evaluated Trojan Attacks}
  \label{tab:attacks}
  \centering
  \begin{tabular}{l|c|c|c}
  \hline
  Properties	                  & BadNets     & Trojaning (SQ)	& Trojaning (WM)\\
  \hline \hline
  \smash{\raisebox{15pt}{\makecell[l]{Example of \\Adv. Input}}}  & \Gape[3pt]{\includegraphics[width=0.5in]{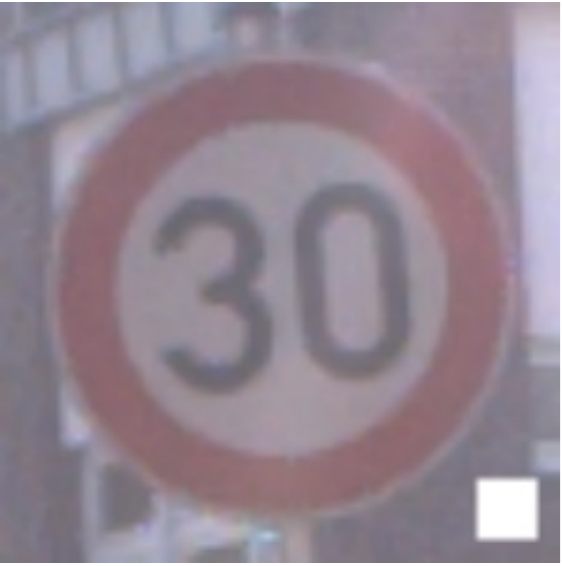}}
                                  & \Gape[3pt]{\includegraphics[width=0.5in]{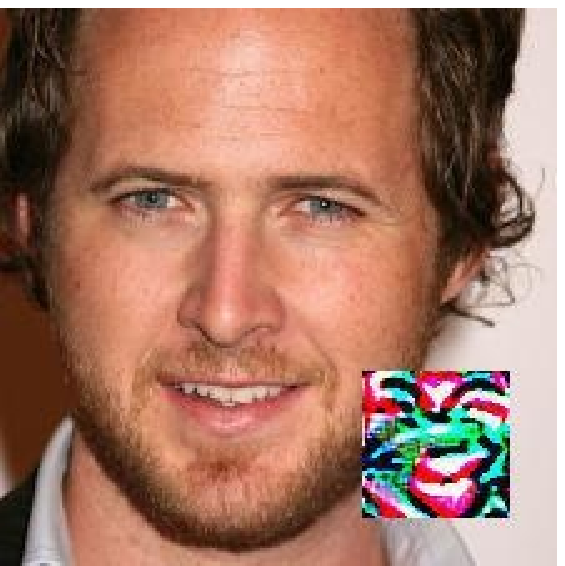}}                   
                                  & \Gape[3pt]{\includegraphics[width=0.5in]{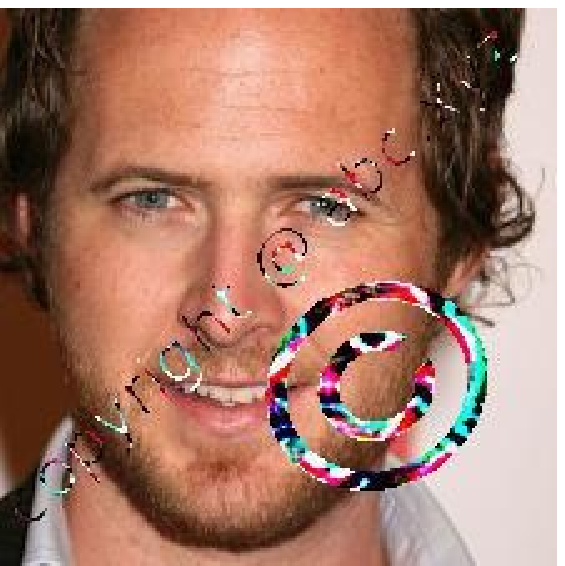}}  	  	\\
  \hline 
  \makecell[l]{Strategy}          & Poisoning   &  Retraining          & Retraining            \\
  \hline
  \makecell[l]{Architecture}      & Resnet34    &  VGG-Face            & VGG-Face              \\
  \hline
  \makecell[l]{Dataset}           & GTSRSB      &  VGG-FAce            & VGG-Face              \\
  \hline
  \makecell[l]{No. Classes}        & 43          &  40			        & 40                    \\
  \hline
  \end{tabular}
\end{table}

\subsection{Evaluation Metrics and Testing Sets}\label{sec:evaluation_metrics}

\textbf{Metrics}. 
The success of a trojan attack can be measured based on two aspects. 
First, the efficiency to keep compromised models having a high accuracy 
(rate of classification to the true class) when processing benign data. 
Second, the attack success rate or ASR, which measures how well triggers 
in adversarial samples activate the misbehavior \cite{liu2017Trojaning}. 
As the latter is expected to be high, trojaned models are also 
characterized by having low accuracy when processing adversarial data.
As a compromised model goes through the healing process our method aims to: 
(1) reduce the ASR to avoid targeted exploits 
(misclassification to the target class),
(2) keep or improve the accuracy when processing benign inputs, and
(3) improve the accuracy with adversarial samples to avoid untargeted exploits
(misclassification to random classes). 
These three factors are the metrics used to measure the performance of
$ConFoc$. 

\textbf{Testing Sets}.
Experiments are run with two versions of a given testing set: 
(1) the given benign version to measure accuracy with benign data and 
(2) its adversarial version to measure accuracy with adversarial data 
and ASR.
The adversarial versions result from adding the trigger to the samples of the
given set.
The size of the sets are given as a percentage of the original training 
set used to create the models.

\subsection{BadNets Attack}\label{sec:evaluation_attack1}

\textbf{Implementation.} 
We conducted the attack against a traffic sign recognition model following
the steps described in \cite{gu2017Badnets}.
The model was built on Resnet34 \cite{he2016Deep} pre-trained 
with the Imagenet dataset \cite{imagenet_cvpr09}.
The pre-trained model was fine-tuned using the 
German Traffic Recognition System Benchmarks (GTRSB) dataset 
\cite{stallkamp2012GTSRB}.

\begin{figure}[!t]
  \centering
  \includegraphics[width=3in]{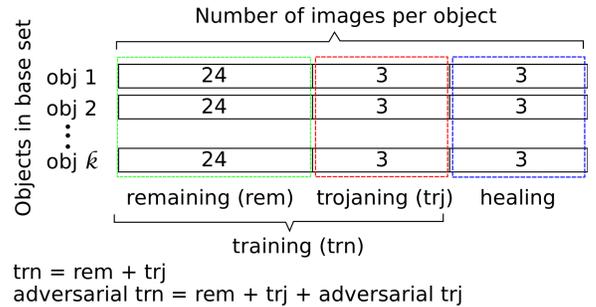}
  \caption{
	Split of the base set (90\% of the GTSRB dataset). 
	\textit{adversarial trj} comprises samples in \textit{trj} with triggers 
	inserted.
  }
  \label{fig:gtsrb_split}
\end{figure}

\textbf{Dataset.} 
GTRSB is a multi-class single-image dataset that contains 39209 
colored training images classified in 43 classes (0 to 42), 
and 12630 labeled testing images. 
The classes are of traffic sign objects such as stop sign, bicycles crossing, and 
speed limit 30 km/h. 
For each physical traffic sign object in the GTRSB training dataset there are 
actually 30 images.
To avoid leakage of information between the data used for training and validation,
the GTRSB training dataset was split by objects in two sets: the validation set 
and the base set.
The validation set was formed by taking 10\% of the objects (30 images per each) 
of every class. 
The other 90\% of objects of each class formed the base set.
The base set was further split to form the final training, trojaning and healing 
sets as shown in Figure \ref{fig:gtsrb_split}.
The split in this case was done based on images.
For each particular object in the base set 3 out 30 images were taken for 
the healing set.
Other exclusive 3 images were taken for the trojaning set (trj), leaving 24 images
per object in the remaining set (rem).
The trojaning set is called adversarial when the chosen trigger is added to its
samples and the samples are mislabeled to the chosen target class.
The training set (trn) comprises both the remaining and trojaning sets 
(trn = rem + trj) and is used in the experiments to create the original 
non-trojaned model $M_O$. 
The adversarial training set, on the other hand, comprises the 
remaining, trojaning, and adversarial trojaning sets 
(adversarial trn = rem + trj + adversarial trj) and is used to train the 
trojaned version of $M_O$ referred to as trojaned model $M_T$.

\textbf{Testing set}. 
Ten images per class were randomly chosen out of the 12630 
available testing samples.

\textbf{Attack strategy.} 
The adversarial samples were generated by inserting a white square as trigger in 
the botton right corner of the benign inputs. 
The size of the square was chosen to be 10\% the size of the smallest between 
the height and width dimensions, and located 5\% of this value from the borders.
Table \ref{tab:attacks}, in row and column 2, shows an adversarial sample with 
the trigger.
Each adversarial sample was labeled with the target class 19, regardless the 
true class of the samples. 

\subsection{Trojaning Attacks: Square (SQ) and Watermark (WM)}
\label{sec:evaluation_attack2}

\textbf{Implementation.} 
For these attacks, $ConFoc$ was tested against two compromised models
provided in \cite{liu2017Trojaning}. 
The two models correspond to the same pre-trained face recognition application 
VGG-Face, infected using two different fine-tuned triggers: square and watermark.
As the accuracy of the provided models was relatively low ($<$ 90\%) when 
tested with the original data in \cite{liu2017Trojaning}, we only consider 
those classes with low rate of misclassification for our experiments. 
The idea was to have an initial trojaned model $M_T$ with high accuracy when 
processing benign data.
To this end, we randomly selected 40 classes including the taget 
class (0 or A.J. Buckley). 
This experimental design choice does not affect the high ASR and low 
accuracy with adversarial samples of the models.

\textbf{Dataset.} 
The original VGG-Face dataset includes $2622000$ images of $2622$ classes 
(1000 images per class).
Currently, no all the images are available and among them there are
a significant amount of mislabeled cases.
Namely, cases in which random images or images of a person A are labeled 
as person B.
For our healing set, we chose 50 out of the available images for each of 
the selected 40 classes.
This represents 5\% of the size of the original dataset.
To reduce the noise caused by the mislabeled cases, only images with frontal 
pose faces were selected.
We then manually cleaned the resulting dataset by removing obvious mislabeled 
samples.
The authors of the attack \cite{liu2017Trojaning} used two sets for testing, 
being one of them extracted from the VGG-Face dataset.
This testing set (referred to by the authors as original dataset) is compound 
of one image per class, and was used to measure the accuracy 
and ASR of the model.
The other testing set was called external dataset and was extracted from 
the LFW dataset \cite{huang2008LFW}.
The images in this set do not necessarily belog to any of the $2622$ classes, 
and were used to measure the ASR only. 
As one of our main goals is to measure the performance of models
with the three metrics listed in Section \ref{sec:evaluation_metrics},
we conducted the experiments with a variation of the original dataset only. 
This ensured a fair comparison with results obtained in previous work.

\begin{table}[!t]
  \caption{Explanation of Acronyms Used in Experiments}
  \label{tab:acronyms}
  \centering
  \begin{tabular}{lp{6.4cm}}
    \toprule
    \makecell[c]{Acronym}  & \makecell[c]{Description} \\
    \midrule
    {$B$}           		& {Set of style base images $\{b^j|j = 1,...,8 \}$ used in
  				       		the $ConFoc$ healing process.}\\  
  				       		    
    {$A$}           		& {Set of style base images $\{a^j|j = 1,...,2 \}$ not used in
  					   		the $ConFoc$ healing process such that $A \cap B = \emptyset$.}\\
  	
  	{$Orig$}             	& {Indicates that the model was evaluated with the original testing  
  	                        set (i.e., without transforming the inputs).}\\ 

  	{$*$}                	& {Indicates that the model was evaluated with styled versions of 
  						  	the testing set (i.e., inputs are transformed).} \\
  
  	{$M_O$}       	   		& {Original non-trojaned model.} \\
   
  	{$M_T$}              	& {Trojaned model.} \\

  	$M_{H(X)}$           	& {Healed model retrained with the retraining set $X_R$. 
                          	$X_R$ is compound of the healing set $X_H$ only.} \\

  	$M_{H(X-0)}$            & {Healed model retrained with the retraining set $X_R$. 
  						  	$X_R$ comprises the healing set $X_H$, and 
  						  	its corresponding content images
  						  	(via Algorithm \ref{alg:content}).} \\

  	$M_{H(X-k)}$            & {Healed model retrained with the retraining set $X_R$.
  						  	$X_R$ is formed by the healing set $X_H$, 
  						  	its content images (via Algorithm \ref{alg:content}), 
  						  	and the styled images generated with the first $k$ 
  						  	style base images in $B$ 
  						  	(via Algorithm \ref{alg:styled}).
  					      	E.g., $M_{H(X-3)}$ means the model is retrained with
  					      	$X_H$, the content images and the styled images generated with
  					      	the style bases $b^{1}$, $b^{2}$, and $b^{3}$ in B.}\\ 				  					   		   
    \bottomrule
  \end{tabular}
\end{table}

\textbf{Testing set}. 
It is formed by 20 random images per class. Two 
adversarial versions of it are used, one for each trigger.

\textbf{Attack strategy.} 
The two provided models were compromised through the retraining 
process covered in Section \ref{sec:background_attacks}.
Row 2 of Table \ref{tab:attacks} shows examples of two adversarial samples 
with the square and watermark triggers in columns 3 and 4 respectively. 
The provided models classify any image with either trigger 
to the target class (0 or A.J. Buckley).

\subsection{Acronyms Used in Experiments}\label{sec:evaluation_acronyms}
Table \ref{tab:acronyms} lists the acronyms used to refer to the models 
and data used in the experiments. 
It also indicates how to identify the testing set used in the evaluation of
each model.

\section{Experiments}\label{sec:experiments}

\begin{figure*}[!t]
\centering
\subfloat[BadNets]{\includegraphics[width=2.2in]{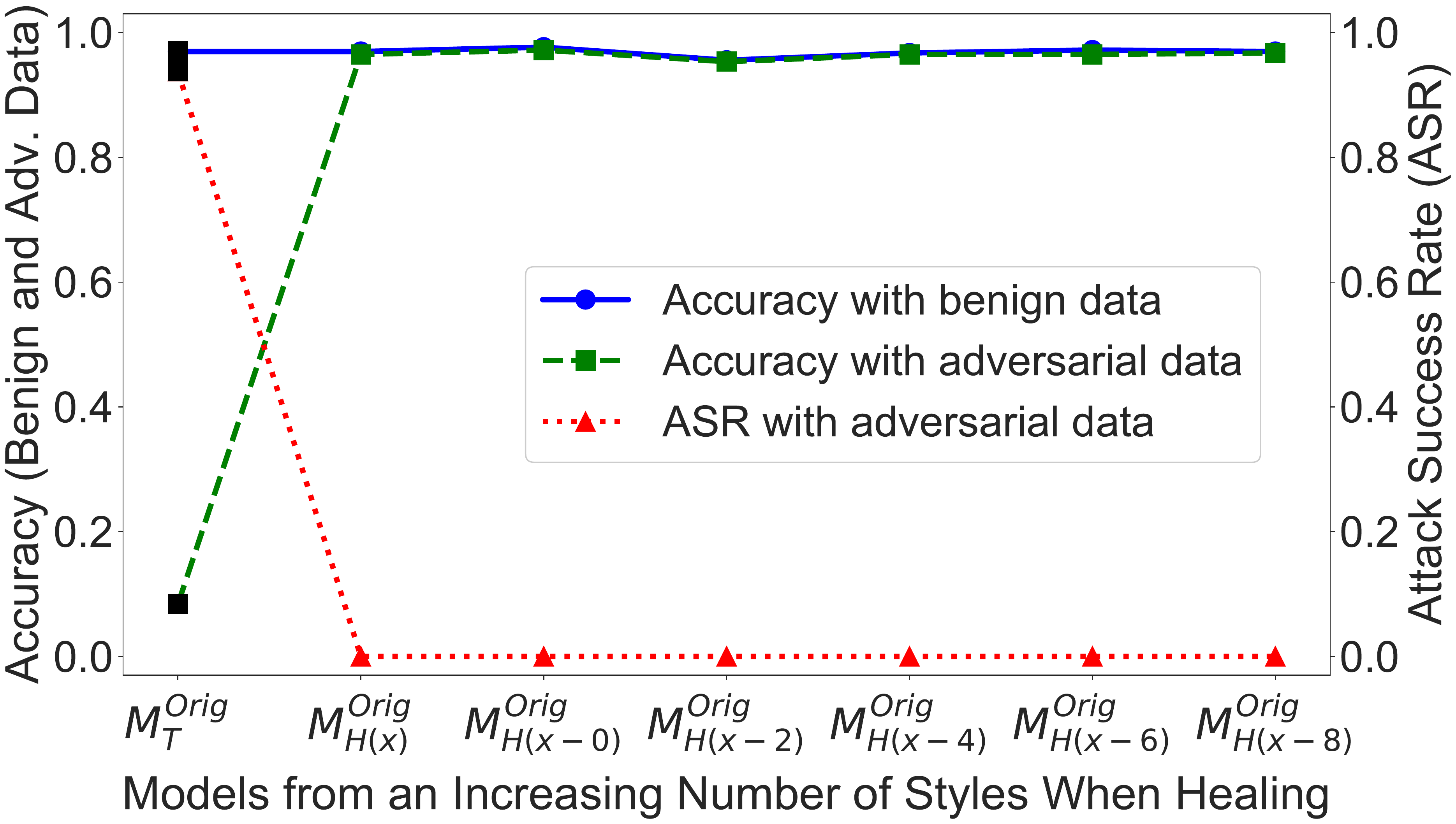}
\label{fig:exp1_badnets}}
\hfil
\subfloat[Trojaning (SQ)]{\includegraphics[width=2.2in]{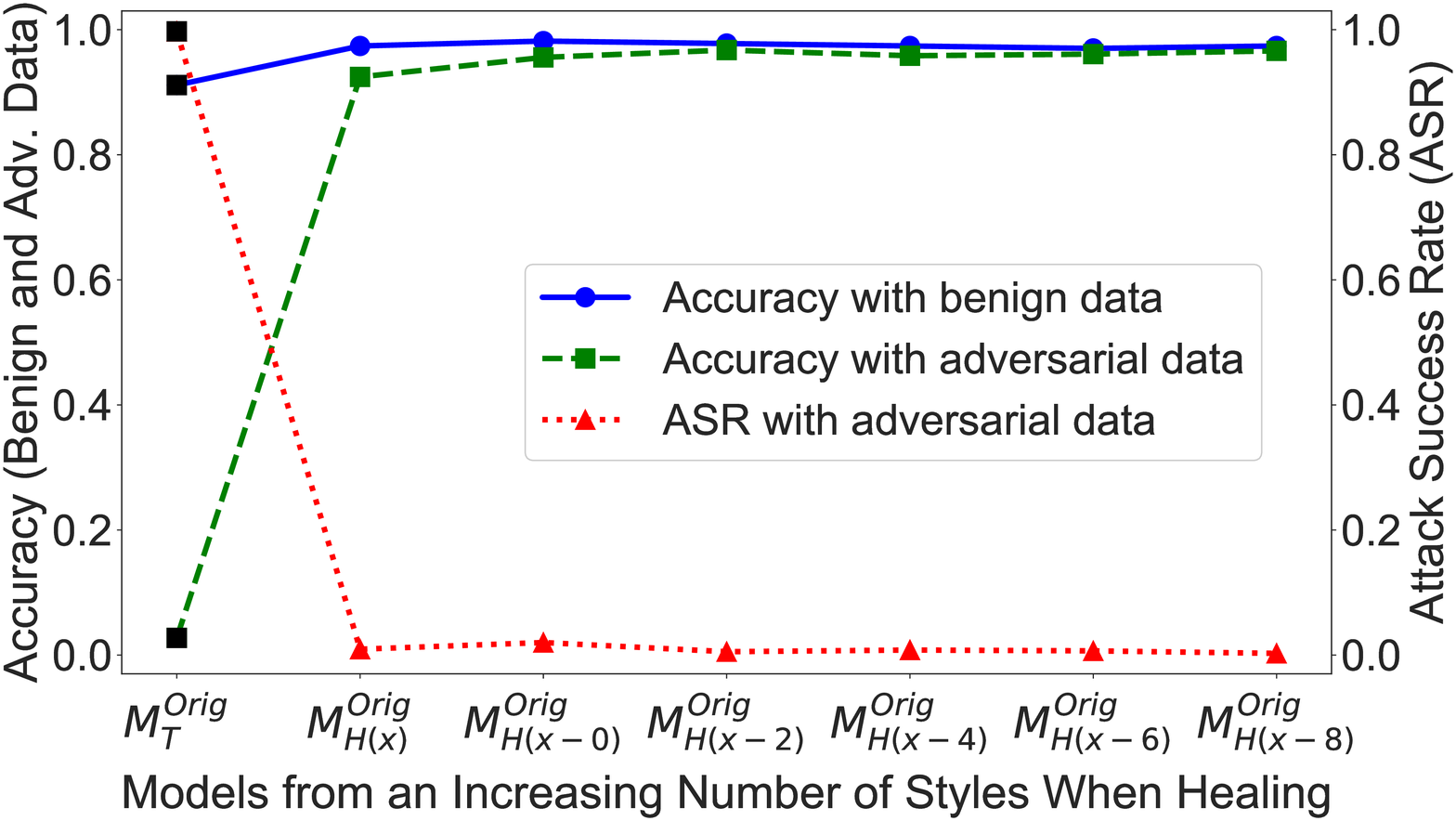}
\label{fig:exp1_trojanSQ}}
\hfil
\subfloat[Trojaning (WM)]{\includegraphics[width=2.2in]{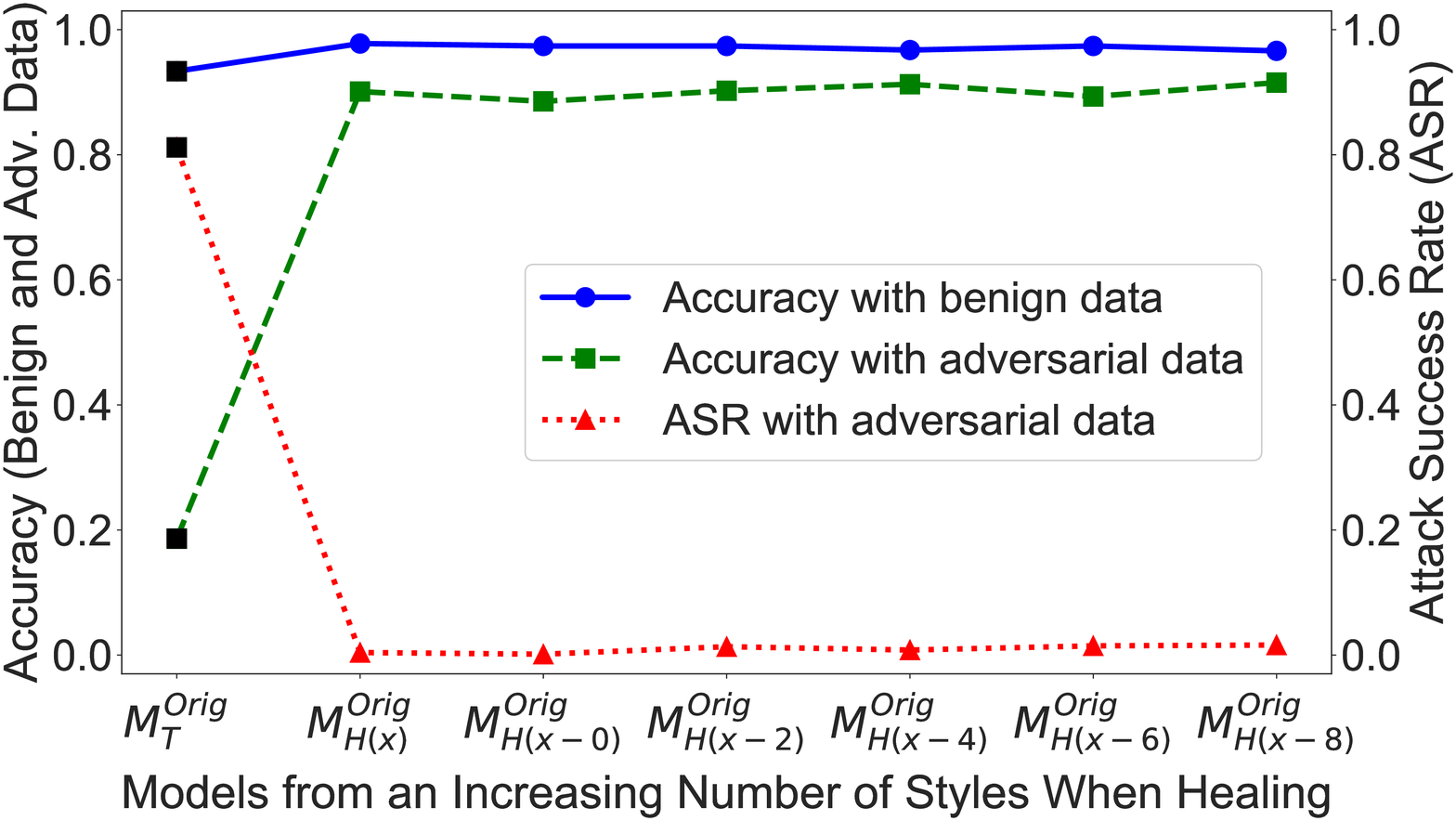}
\label{fig:exp1_trojanWM}}
\caption{
 Metric variations as \textit{ConFoc} is progressively applied to $M_T$ 
 by increasing the number styles used in the healing process.
 Resulting healed models (in $x$-axis) are evaluated with the original 
 (non-transformed) testing datasets (refer to Table \ref{tab:acronyms}).
}
\label{fig:exp1_original_inputs}
\end{figure*}

This section describes the experiments conducted to evaluate
$ConFoc$ against trojan attacks.
The experiments were designed to answer a series of research questions,
included in each of the following subsections along with our findings.

\subsection{Robustness When Processing Original Inputs}
\label{sec:experiments_exp1}
    \textbf{RQ1.} How do the evaluation metrics change as $ConFoc$ is 
    progressively applied using an incremental number of styles? 

We investigate whether the performance of trojaned models (based on 
the three metrics described in Section \ref{sec:evaluation_metrics}) 
improve as we increase the number of styles used in the healing 
process.
We conduct our evaluation against BadNets, Trojaning (SQ), and 
Trojaning (WM) using the corresponding original testing sets.
Figure \ref{fig:exp1_original_inputs} shows the results.
For each of the attacks, we start with the corresponding trojaned 
model $M_T$ and proceed as follows.
	First, $M_T$  is evaluated with the original testing samples to measure 
	the performance of the model before applying $ConFoc$ 
	($M_T^{Orig}$ in $x$-axis). 
	The corresponding points in the plots are marked with a black 
	square to highlight that these are the initial values of the 
	metrics.
	Then, $M_T$ is taken through the $ConFoc$ healing process
	multiple times using incremental retraining sets to measure 
	how the metrics vary as more styles are used  
	(points $M_{H(X)}^{Orig}$ to $M_{H(X-8)}^{Orig}$ in $x$-axis).

Figures \ref{fig:exp1_badnets}, \ref{fig:exp1_trojanSQ}, and 
\ref{fig:exp1_trojanWM}
show that the performance improves as $ConFoc$ is progressively
applied.
The three metrics tend to converge to the aimed values with just a 
few styles (considering the graphs of all the metrics, two styles 
suffice for all the cases).
For the three attacks, the ASR drops to or close to 
0.0\%.
Simultaneously, the accuracy with benign data converges to high values 
that outperform the initial accuracy of the trojaned model. 
This metric has percentage increases of 0.24\%, 7.28\%, and 3.63\% in the 
best obtained healed models
$M_{H(X-6)}$, $M_{H(X-4)}$, and $M_{H(X-4)}$ 
for the attacks BadNets, Trojaning (SQ), and Trojaning (WM) 
respectively.
For the accuracy with adversarial data, we also obtain a significant 
increase in these models. 
This accuracy increases 88.14\%, 94.02\%, and 72.66\% in the models 
for the same order of attacks.
An interesting behavior is observed in the case of Trojaning (WM). 
The accuracy with adversarial data significantly improves to values 
above 90\% in all the cases, 
but always remains lower than the accuracy achieved with benign data. 
This phenomenon can be explained by the fact that the watermark overrides 
the object of interest (i.e., faces), covering certain key attributes of the 
faces (e.g., eyes, lips, etc.) used by models during the classification
(a violation to the condition $C3$ listed in Section 
\ref{sec:background_attack_definition}).
As a consequence, some adversarial inputs with the watermark covering key 
attributes of the faces cannot be recognized to their true classes after 
applying $ConFoc$ because the resulting contents (face shapes plus watermark) 
are not part of the content of images present in the healing set $X_H$.
Note that a violation to condition $C3$ means that attackers assume 
weak defenders who cannot perceive triggers even when they cover a 
significant portion of the input images (a less real-world feasible 
scenario from the standpoint of the attacker). 

\noindent
\fbox {
  \parbox{8.5cm}{
    \textbf{Findings.} With a few styles (two in our case), 
    $ConFoc$ reduces the ASR to or close to 0.00\%, while ensures 
    that both accuracies converge to close values equal or above 
    the original accuracy when conditions $C1$-$C4$ are satisfied.
  }
}
 
\subsection{Robustness When Processing Transformed Inputs}
\label{sec:experiments_exp2}
    \textbf{RQ2.} How well do models learn to focus on content 
    			  and how effective $ConFoc$ is when processing 
    			  styled inputs?   

\begin{figure*}[!t]
\centering
\subfloat[BadNets: Accuracy (Benign Data)]{\includegraphics[width=2.2in]{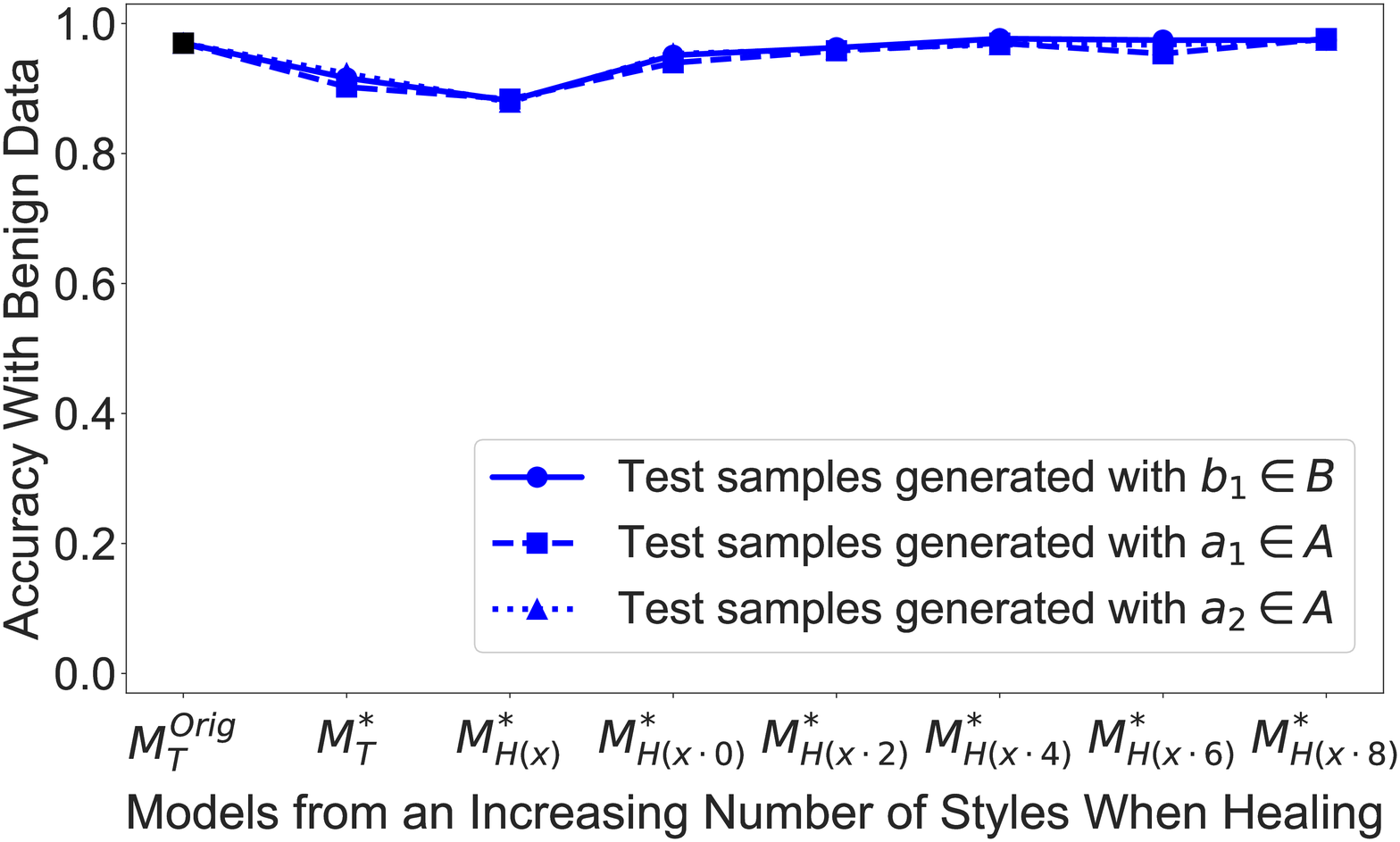}
\label{fig:exp2_acc_ben_badnets}}
\hfil
\subfloat[BadNets: Accuracy (Adversarial Data)]{\includegraphics[width=2.2in]{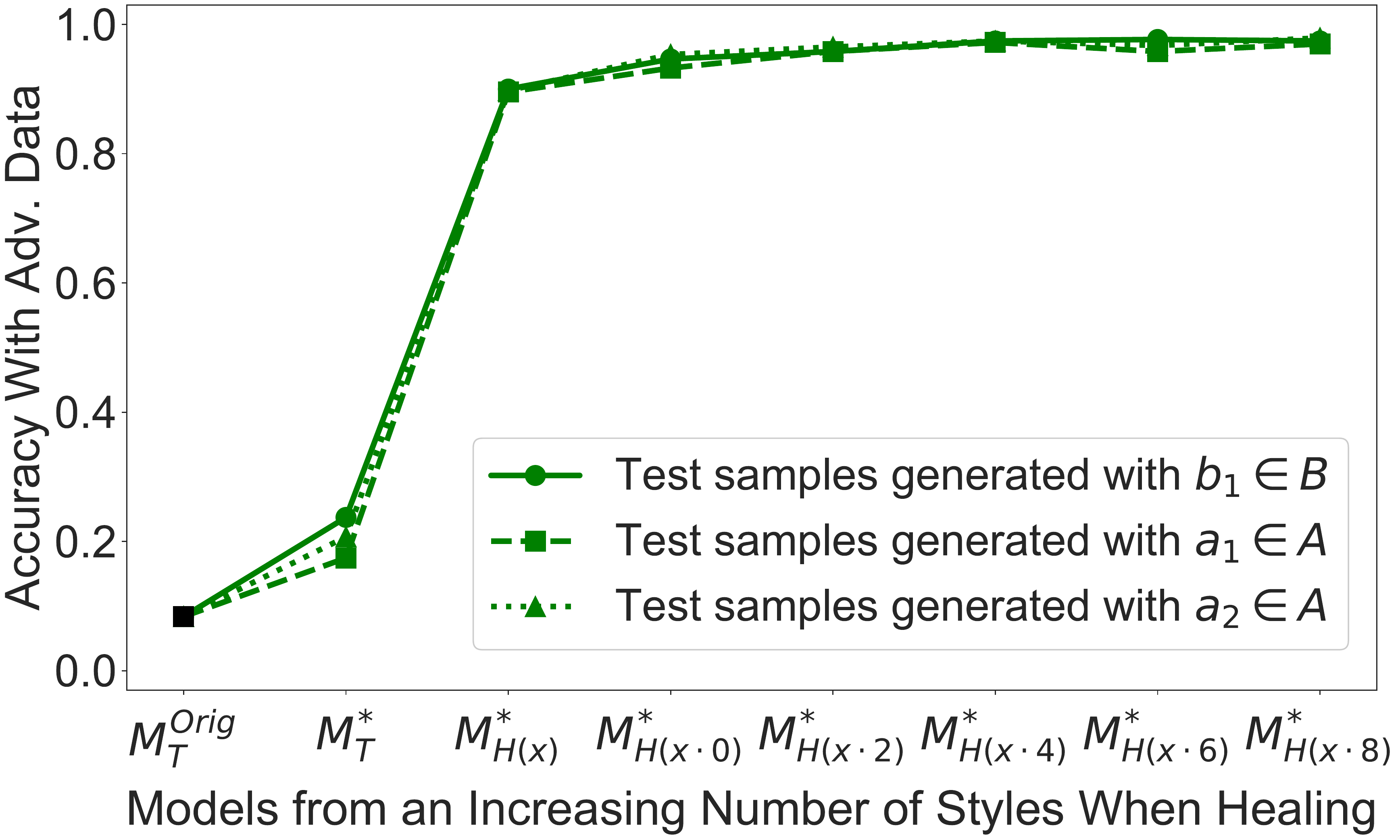}
\label{fig:exp2_acc_adv_badnets}}
\hfil
\subfloat[BadNets: ASR (Adversarial Data)]{\includegraphics[width=2.2in]{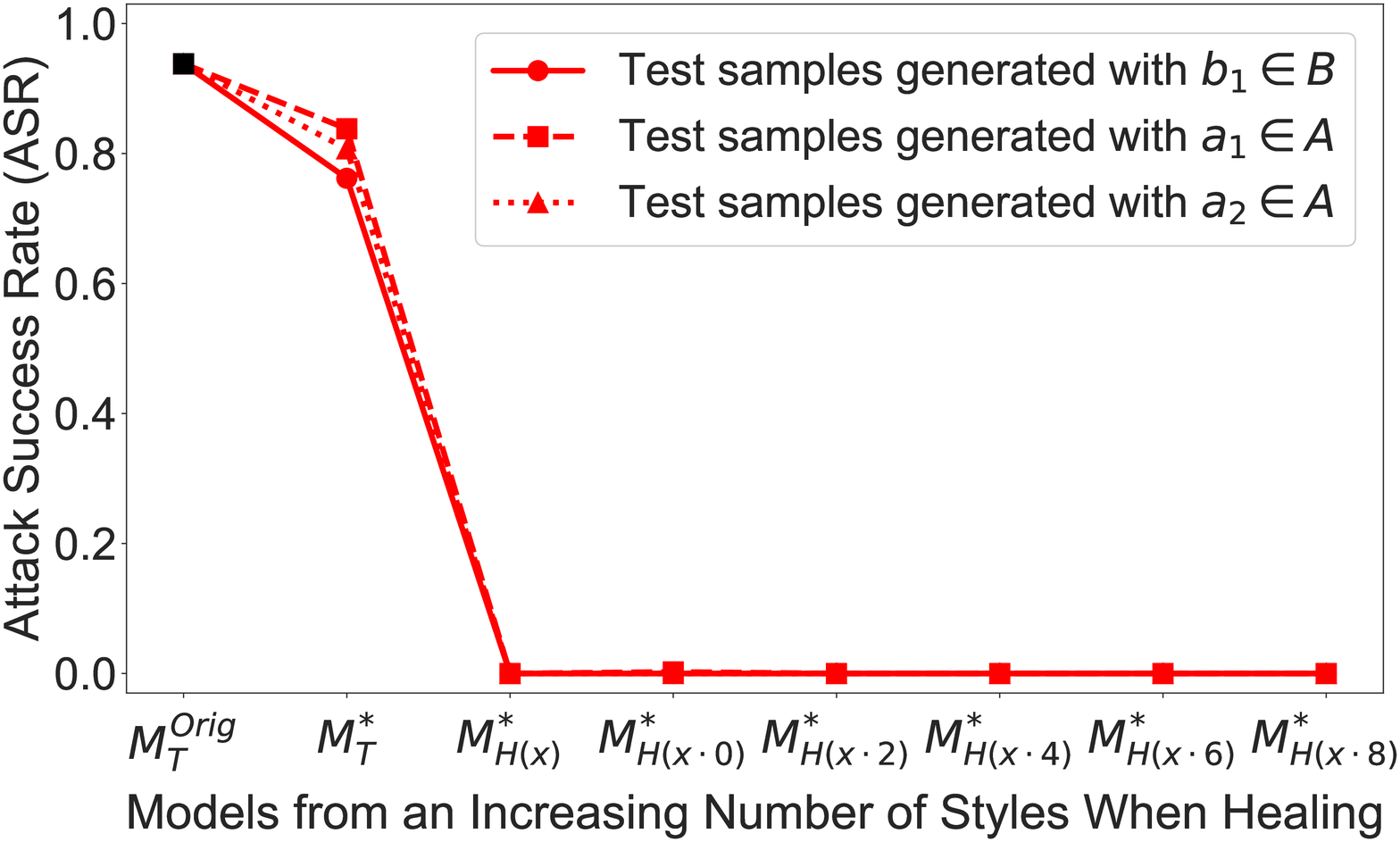}
\label{fig:exp2_asr_adv_badnets}}
\hfil 
\subfloat[Trojaning (SQ): Accuracy (Benign Data)]{\includegraphics[width=2.2in]{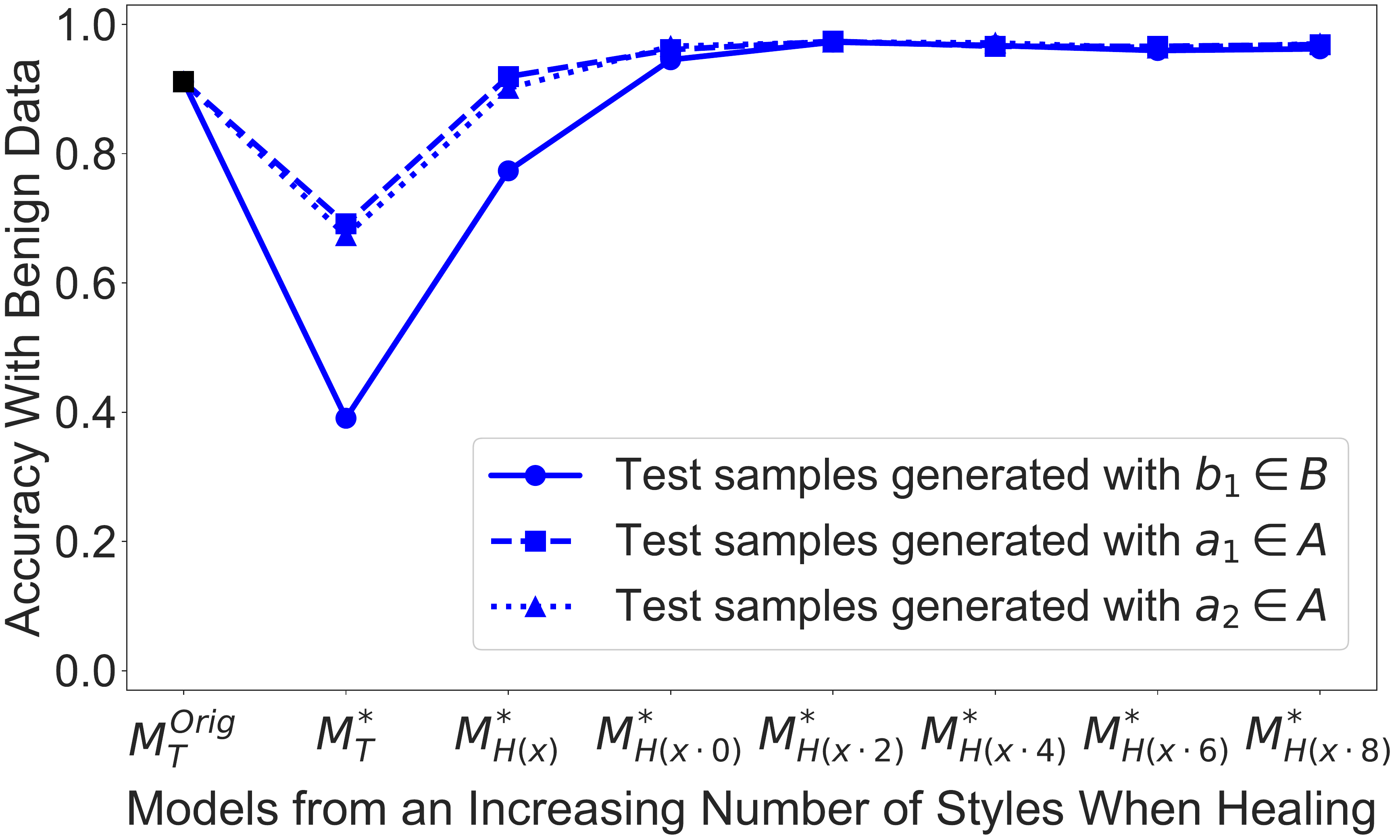}
\label{fig:exp2_acc_ben_trojanSQ}}
\hfil
\subfloat[Trojaning (SQ): Accuracy (Adversarial Data)]{\includegraphics[width=2.2in]{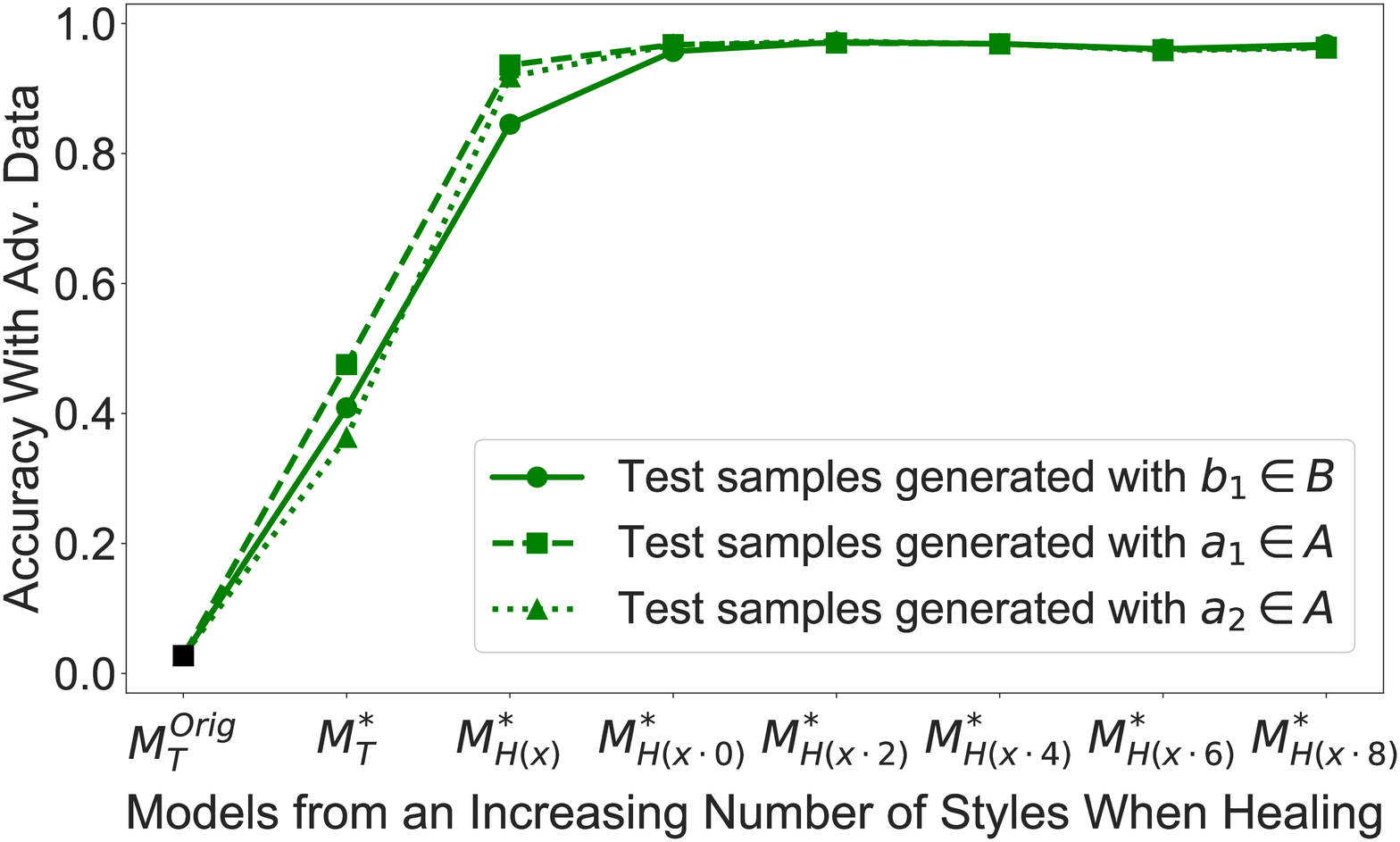}
\label{fig:exp2_acc_adv_trojanSQ}}
\hfil
\subfloat[Trojaning (SQ): ASR (Adversarial Data)]{\includegraphics[width=2.2in]{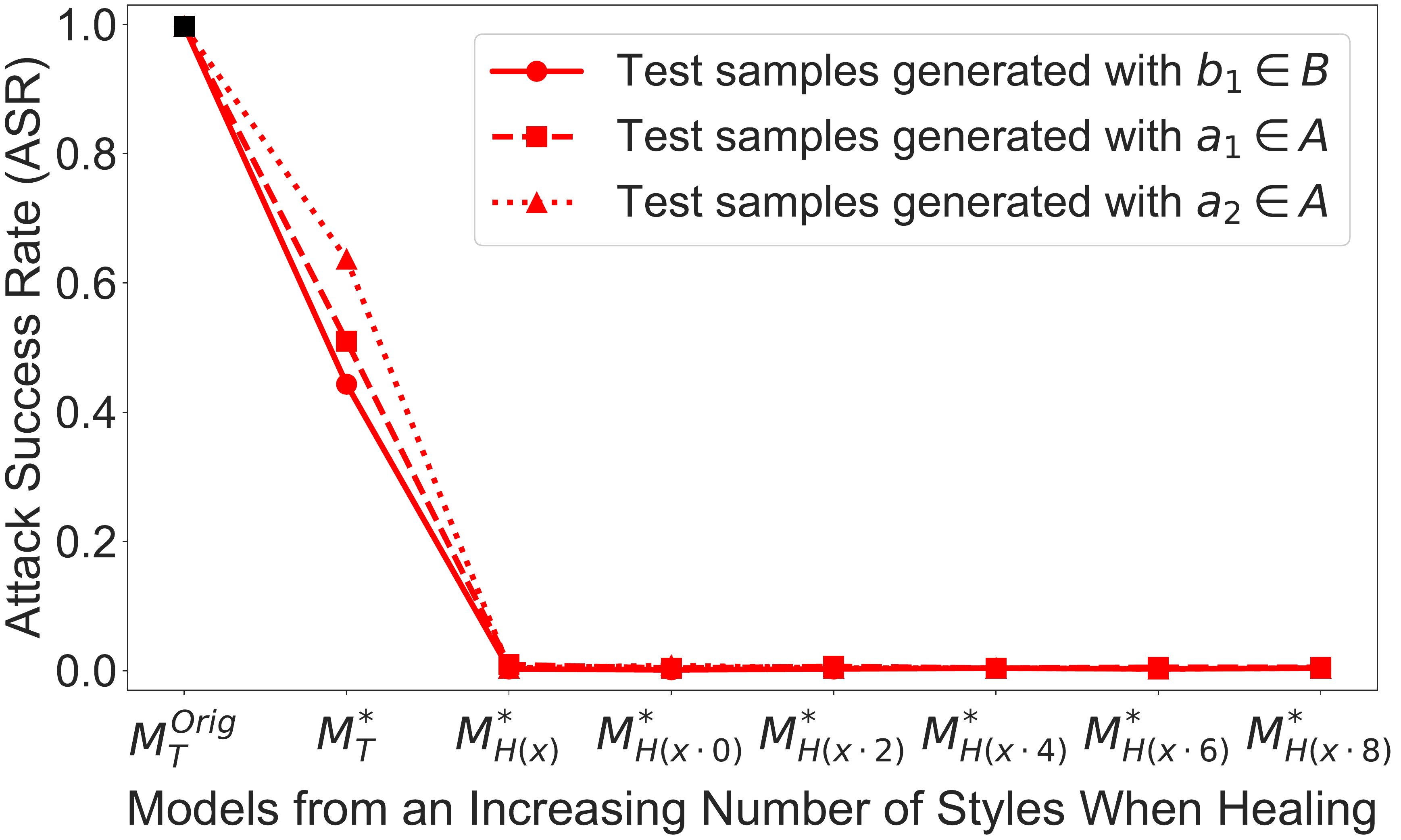}
\label{fig:exp2_asr_adv_trojanSQ}}
\hfil
\subfloat[Trojaning (WM): Accuracy (Benign Data)]{\includegraphics[width=2.2in]{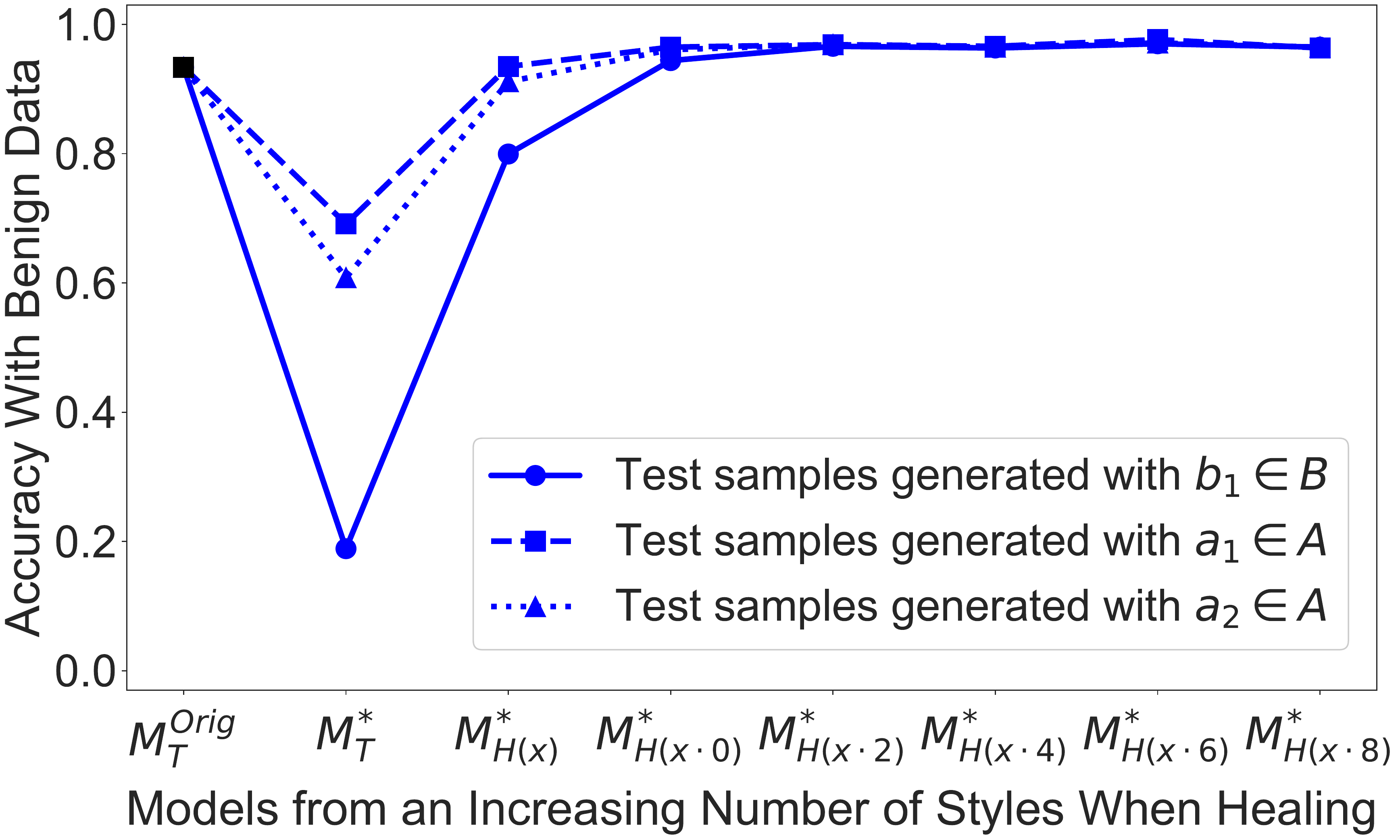}
\label{fig:exp2_acc_ben_trojanWM}}
\hfil
\subfloat[Trojaning (WM): Accuracy (Adversarial Data)]{\includegraphics[width=2.2in]{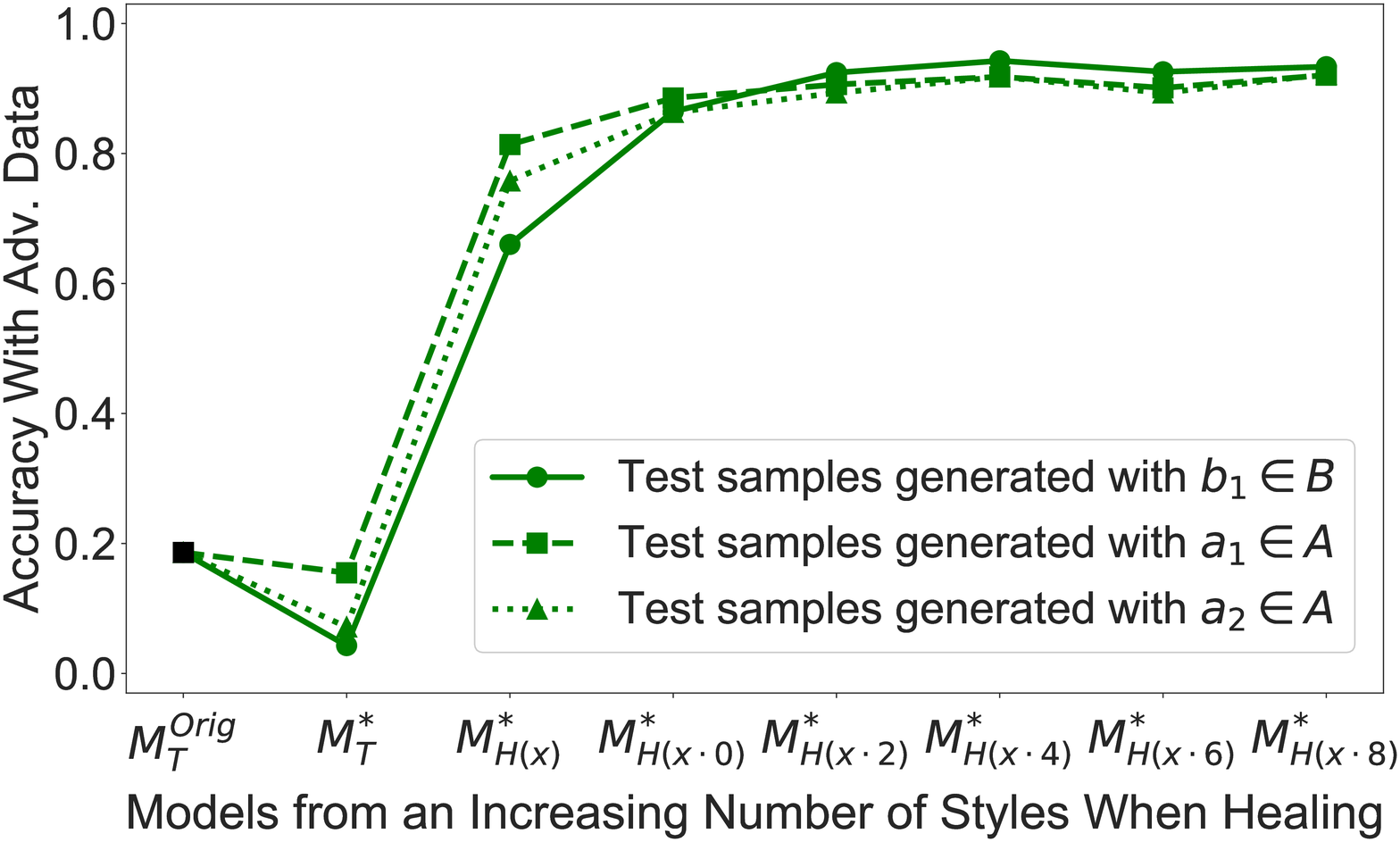}
\label{fig:exp2_acc_adv_trojanWM}}
\hfil
\subfloat[Trojaning (WM): ASR (Adversarial Data)]{\includegraphics[width=2.2in]{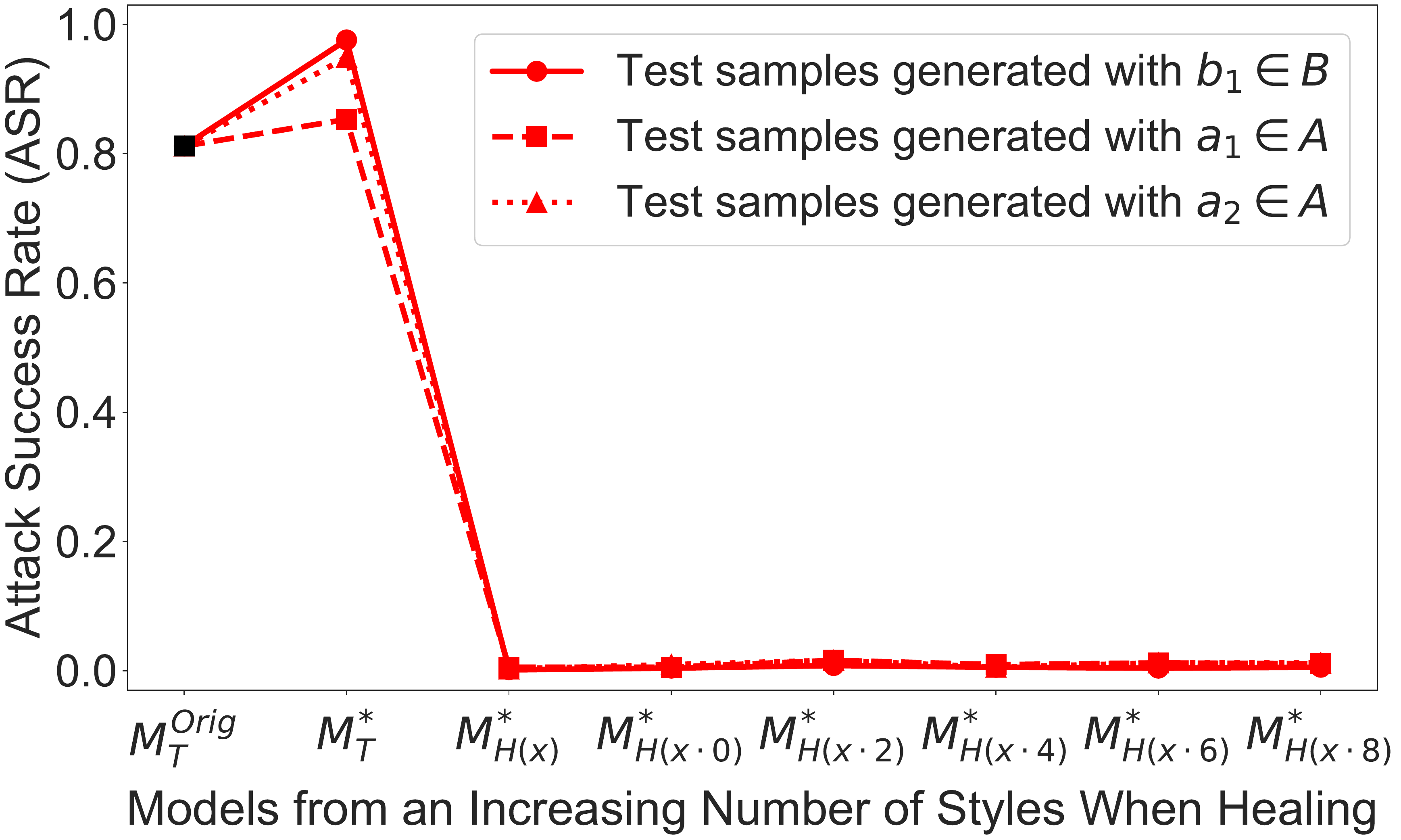}
\label{fig:exp2_asr_adv_trojanWM}}
\caption{
 Efficiency of $ConFoc$ on making models focus on content at testing.
 $M_T$ is healed with an incremental number of styles. 
 Resulting healed models are evaluated with test sets generated with 
 different style bases: $b^1 \in B$, $a^1 \in A$, and $a^2 \in A$. 
}
\label{fig:exp2_transformed_inputs}
\end{figure*}

Following the methodology of the previous section,
we now evaluate how well healed models learn to focus on 
the content of images, disregarding their styles.
To this end, models are evaluated using three different styled or 
transformed versions of the testing set.
One version is generated with the style base image $b^1 \in B$, 
which is used in the healing process. 
The other two versions are obtained using the style base images 
$a^1$ and $a^2$ in $A$, which are not used during the healing 
of the models.
%
For each attack, we start again with the corresponding trojaned model 
$M_T$ evaluated with original samples to get the initial values of the 
metrics before applying $ConFoc$ ($M_T^{Orig}$ in $x$-axis).
Following, $M_T$ is tested using transformed samples to measure 
the impact that the input transformation itself has on the performance 
($M_T^*$ in $x$-axis).
Finally, transformed samples are used to test the models healed through 
an incremental application of $ConFoc$ 
(points $M_{H(X)}^*$ to $M_{H(X-8)}^*$ in $x$-axis).

Figure \ref{fig:exp2_transformed_inputs} shows the results. 
Each subfigure in it corresponds to one of the metrics and an attack.
For all the metrics, the final performance of the healed models are nearly 
the same, regardless the styled version of the testing set used in the 
evaluation.
The metrics tend to converge to sought values as more styles are used.
This is a consequence of the increasing data augmentation achieved 
through the addition of new styles to the $ConFoc$ process.
Both accuracies improve because the larger the retraining set is, 
the more samples with common content information the model receives.
With the increasing sets, models are fine-tuned with enough samples for
them to extract the contents of the training sample features, 
which are also present in the testing samples.
Simultaneously, the attack success rate also drops because of this data 
augmentation.
As the retraining set increases, models tend to forget the trigger 
because more parameter updates are executed in one epoch of 
training with samples not including the trigger.
This is an expected behavior based on the findings of Liu et al. 
\cite{liu2017Neural}, who shows that this metric decreases as 
more benign data samples (original version only) are used to 
fine-tune DNN models.

Notice that the transformed testing datasets used in this evaluation are 
generated with both styles used and not used in the healing process.
Hence, this experiment shows the effectiveness of $ConFoc$ on making
models focus on content and not on styles during the classification. 
One interesting observation is that using styled images without healing 
the models does not prevent the attacks.
The attacks become ineffective after applying $ConFoc$ with a few 
styles. Considering all the plots and metrics in Figure 
\ref{fig:exp2_transformed_inputs}, four styles suffice.

After $ConFoc$, the ASR is reduced to or close to 0.0\%.
In all the attacks, the accuracies with benign data (regardless the style) 
achieve high values that outperform the initial accuracies of the trojaned
model.
Using the best resulting healed models 
$M_{H(X-6)}$, $M_{H(X-4)}$ and $M_{H(X-4)}$ 
for the attacks BadNets, Trojaning (SQ), and Trojaning (WM) respectively, 
this metric grows 0.47\%, 6.71\%, and 3.2\% when evaluated with the 
transformed testing set generated with $b_1 \in B$.
With respect to the accuracy with adversarial data, the metric increases
89.30\%, 94.41\%, and 75.65\% with the same healed models.

\noindent
\fbox {
  \parbox{8.5cm}{
    \textbf{Findings.} With a few styles (four in our case), 
    $ConFoc$ reduces the ASR to or close to 0.00\%, while ensures 
    both accuracies get values equal or above the original accuracy 
    regardless the input style when conditions $C1$-$C4$ hold.
  }
}

\subsection{Effect on Non-Trojaned Models} 
\label{sec:experiments_exp3}
    \textbf{RQ3.} What is the impact of $ConFoc$ on the accuracy 
    (only benign data applies) of non-trojaned models?   

One of the main challenges defending against trojan attacks is the 
lack of tools to determine whether a given model has a trojan.
Due to this restriction, this section evaluates the impact $ConFoc$ has on 
the accuracy of an original non-trojaned model $M_O$.
Our goal is determining whether $ConFoc$ can be applied
to any model (whether infected or not) without impairing its current 
performance (accuracy with benign data).

\begin{figure}[!t]
  \centering
  \includegraphics[width=3.0in]{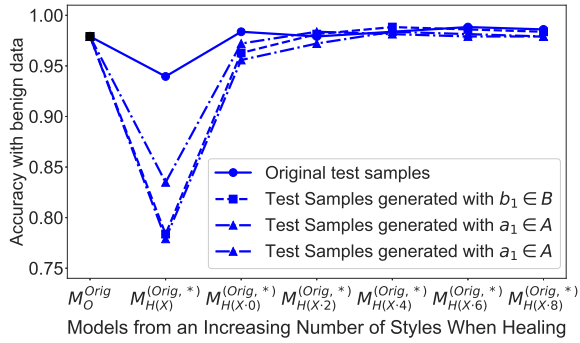}
  \caption{
	Accuracy (benign data) variation of the GTSRB-based non-trojaned 
	model when \textit{ConFoc} is progressively applied. 
   }
  \label{fig:exp3_acc_ben_badnets}
 \end{figure}

We take the non-trojaned version of the models created with the datasets
GTSRB and VGG-Face through the $ConFoc$ healing process.
Figure \ref{fig:exp3_acc_ben_badnets} shows the metric variation of the 
GTSRB model as styles are added to the healing process.
Taking model $M_{H(X-4)}$ tested with the transformed samples generated with 
$b_1 \in B$ as example, 
the accuracy improves from 97.91\% to 98.37\%.
We get a similar graph (not included) with the VGG-Face model. 
In this case, the best performance is obtained with the model $M_{H(X-6)}$, 
with a percentage increase of 0.17\%. 
These results prove that $ConFoc$ does not affect the accuracy 
of non-trojaned models. 
In contrast, the trends of the graphs shows that it at least remains 
the same if enough styles are used in the healing process.

\noindent
\fbox {
  \parbox{8.5cm}{
    \textbf{Findings.} $ConFoc$ can be equally applied to any model
    (either trojaned or not) as it does not impair its performance.  
  }
}

\subsection{Healing Set Size and Number of Styles}
\label{sec:experiments_exp4}
    \textbf{RQ4.} Does the number of styles required in the $ConFoc$ 
    healing process depend on the size of the healing set $X_H$? 

We investigate the relationship between the number of styles 
required to successfully apply $ConFoc$ and the size of the 
healing set.
%
This is a key question because having access to extra 
training sets is challenging in real-world scenarios.
As specified in Section \ref{sec:evaluation}, previous experiments
are run with healing sets of size 10\% and 5\% 
for the models infected with BandNets and Trojaning Attack 
respectively. 
We now replicate the same experiments progressively 
decreasing the size of these sets and selecting the model with
the best performance in each case.
Table \ref{tab:number_of_styles} shows that there is no relationship
between size of the healing set and the number of styles needed to 
apply $ConFoc$.
This can be explained because the combination of some contents and 
styles add noise to the resulting retraining set, which make models 
to not monotonically improve as more styles are added.
Whereby, defenders need to apply the best training practices to fine-tune
the models with the generated data so as to obtain the best possible results.
%

\noindent
\fbox {
  \parbox{8.5cm}{
    \textbf{Findings.} There is no relationship between the $X_H$ 
    size and the number of styles needed to successfully apply $ConFoc$. 
  }
}

\subsection{Performance and Comparison With the State-of-the-Art}
\label{sec:experiments_exp5}
    \textbf{RQ5.} How well does $ConFoc$ perform compared to the 
    state-of-the-art and what overhead it imposes at testing? 

Table \ref{tab:comparison} shows the performance of $ConFoc$ and its 
comparison with the state-of-art Neural Cleanse 
\cite{wang2019NeuralCleanse}.
To be complete in our comparison with fine-tuning-based methods, we 
also include a comparison with Retraining \cite{liu2017Neural}.
The table contains the accuracies (with both begin and adversarial data) 
and the ASR after applying the defensive methods.
The first column specifies the attack used for the evaluation.
DS refers to the size of the healing set.
The initial values of the trojaned models (before applying any 
method) are included in Table \ref{tab:initial}. 

In Table \ref{tab:comparison},
columns below ConFoc (Original Inputs) and ConFoc (Transformed Inputs) 
summarize the performance of our method using the best healed 
models (specified in Table \ref{tab:number_of_styles}) for different sizes 
of the healing set.
As the names indicate, we tested $ConFoc$ with both original and transformed 
inputs.
The other two techniques (Retraining and Neural Cleanse) were evaluated with 
original inputs as the methods require.
With Retraining, we fine-tuned the model using the original healing set 
for multiple epochs and selected the best resulting model.
In the case of Neural Cleanse, we proceeded exactly as indicated by the authors
in \cite{wang2019NeuralCleanse}.
We added the reversed-engineered triggers to 20\% of the samples in the healing 
set and retrained the model for one epoch.
The reversed-engineered triggers are provided by the authors in 
\cite{wang2019NCRepo}.
During the execution of this method, only the trigger related to 
Trojaning (WM) worked as expected. 
Whereby, we ran the Neural Cleanse method against BadNets and 
Trojaning (SQ) with the actual triggers used to conduct the 
attacks.
This action does affect the performance of Neural Cleanse.
In contrast, it represents the ideal scenario in which triggers 
are perfectly reverse-engineered and the produced corrected models 
provide the best possible results.

\begin{table}[!t]
  \caption{Best Healed Models After Applying ConFoc}
  \label{tab:number_of_styles}
  \centering
  \begin{tabular}{lccc}
    \toprule   
      	  \multicolumn{2}{c}{Experimental Setup} 
    	& \multicolumn{2}{c}{Best Healed Model}\\
    \cmidrule(r){1-2}
    \cmidrule(r){3-4}   			
    Attack           & DS  		& Model ID		& No. Styles (Including Content) 	\\		
    \midrule
    BadNets   		& 10\% 		& $M_{H(X-6)}$	& 7   								\\	   
    Trojaning (SQ)  & 5\%		& $M_{H(X-2)}$	& 3									\\  
    Trojaning (WM)  & 5\%		& $M_{H(X-4)}$	& 5									\\
    \midrule
    BadNets   		& 6.6\% 	& $M_{H(X-4)}$	& 5   								\\	   
    Trojaning (SQ)  & 3.3\%		& $M_{H(X-2)}$	& 3									\\  
    Trojaning (WM)  & 3.3\%		& $M_{H(X-1)}$	& 2									\\
    \midrule
    BadNets   		& 3.3\% 	& $M_{H(X-4)}$	& 5   								\\	   
    Trojaning (SQ)  & 1.67\%	& $M_{H(X-1)}$	& 2									\\  
    Trojaning (WM)  & 1.67\%	& $M_{H(X-2)}$	& 3									\\    	   	 
    \bottomrule
  \end{tabular}
\end{table}

\newcolumntype{g}{>{\columncolor{Lightgrey}}c} 
\begin{table*}[!t]
  \caption{Metrics of Corrected Models After Applying ConFoc and 
           Other State-of-the-Art Model Hardening Techniques
          }
  \label{tab:comparison}
  \centering
  \scalebox{0.95}{
  \begin{tabular}{lccgccgccgccgc}
    \toprule
      	  \multicolumn{2}{c}{Experimental Setup}
    	& \multicolumn{3}{c}{ConFoc (Original Inputs)} 
    	& \multicolumn{3}{c}{ConFoc (Transformed Inputs)} 
    	& \multicolumn{3}{c}{Retraining (Healing Set Only)} 
    	& \multicolumn{3}{c}{Neural Cleanse}\\
    \cmidrule(r){1-2}
    \cmidrule(r){3-5}
    \cmidrule(r){6-8}
    \cmidrule(r){9-11}
    \cmidrule(r){12-14}
    \rowcolor{White}
    Attack          & DS		& Acc (Ben) &Acc (Adv)		& ASR	    	& Acc (Ben) &Acc (Adv)		& ASR			& Acc (Ben) &Acc (Adv)		& ASR    		& Acc (Ben) &Acc (Adv)		& ASR  	   		\\
    \midrule
    BadNets   		& 10.0\%  	& 97.21\% 	& 96.51\%  		& 0.00\%	    & 97.44\% 	& 97.67\%  		& 0.00\%    	& 96.98\% 	& 96.51\%  		& 0.00\% 		& 97.21\%    & \rc 65.35\%  & 0.00\%   		\\   
    Trojaning (SQ)  & 5.0\% 	& 97.79\%	& 96.75\%		& 0.53\%	    & 97.27\% 	& 97.14\% 		& 0.27\%    	& 97.40\%	& 92.45\%		& 0.94\% 		& 97.53\%    & 97.27\% 		& 0.27\%   		\\
    Trojaning (WM)  & 5.0\% 	& 96.75\%	& \dc91.28\%	& 0.80\%	    & 96.35\%	& \dc94.27\%	& 0.53\%    	& 97.79\%	& 90.10\%		& 0.40\%		& 96.88\%    & 93.36\% 		& 0.27\%   		\\ 	
    \midrule
    BadNets   		& 6.66\%  	& 97.21\% 	& 96.98\%  		& 0.00\%	    & 97.21\% 	& 97.21\%  		& 0.00\%    	& 96.28\% 	& 94.42\%  		& 0.00\% 		& 97.21\%    & \rc 54.42\%  & 0.00\%		\\
    Trojaning (SQ)  & 3.33\% 	& 97.40\%	& 97.27\%		& 0.53\%	    & 96.75\%	& 97.14\%		& 0.66\%   	    & 98.31\%	& \rc 82.68\%	& 0.16\% 		& 97.53\%    & 97.79\% 		& 0.13\%		\\
    Trojaning (WM)  & 3.33\% 	& 96.88\%	& \dc92.32\%	& 0.13\%	    & 96.09\%	& \dc91.67\%	& 0.27\%   	    & 98.05\%	& 92.19\%	    & \rc 1.73\%	& 97.01\%    & 92.45\% 		& 0.00\%    	\\	
    \midrule
    BadNets   		& 3.33\%  	& 96.05\% 	& 96.05\%  		& 0.00\%	    & 97.21\% 	& 97.21\%  		& 0.00\%    	& 95.12\% 	& 96.05\%  		& 0.00\% 		& 97.21\%,   & \rc 58.84\%  & \rc 3.33\%	\\
    Trojaning (SQ)  & 1.67\% 	& 98.05\%	& 96.09\%		& 0.67\%	    & 96.35\%	& 96.88\%		& 0.27\%    	& 98.05\%	& \rc 82.94\%	& 0.16\%		& 96.61\%    & 96.09\% 		& 0.00\%		\\
    Trojaning (WM)  & 1.67\% 	& 97.27\%	& \dc91.15\%	& 0.13\%	    & 95.96\%	& \dc92.84\%	& 0.27\%  	    & 97.66\%	& \rc 83.72\%	& \rc 9.08\% 	& 96.48\%    & \rc 89.58\% 	& \rc 1.34\%    \\	
    \bottomrule
  \end{tabular}
  }
\end{table*}

\begin{table}[!t]
  \caption{Initial Metrics of Trojaned Models}
  \label{tab:initial}
  \centering
  \begin{tabular}{lccc}
    \toprule      			
    Attack          & Acc (Ben) &Acc (Adv)	& ASR	    \\		
    \midrule
    BadNets   		& 96.98\% 	& 8.37\%	& 93.81\%   \\	   
    Trojaning (SQ)  & 91.15\%	& 2.73\%	& 99.73\%	\\  
    Trojaning (WM)  & 93.36\%	& 18.62\%	& 81.18\%	\\	   	 
    \bottomrule
  \end{tabular}
\end{table}

As shown in the Table \ref{tab:comparison}, all the defensive methods 
produce high accuracy with benign data regardless the size of the 
healing set. In most cases, this metric is superior to the initial 
value of the trojaned model (see to Table \ref{tab:initial}). 
The main differences between the methods are observed in the 
accuracy with adversarial data (highlighted in light grey for 
all the methods) and the ASR.
$ConFoc$ (with both original and transformed inputs) 
constantly gets high values in these two metrics, while the other 
methods produce values below 90\% for the former and above 1\% 
for the latter as the healing set decreases.
These cases are marked in red in the table.

With respect to the accuracy with adversarial data, 
Retraining, as expected, tends to produce models with lower 
values in this metric as the healing sets become smaller in 
all the attacks \cite{liu2017Neural}.
Neural Cleanse produces models that perform well against both 
Trojaning Attacks and unwell against BadNets regardless the 
size of the healing set.
This is because Neural Cleanse relies on updating the model  
parameters for one epoch only, which does not suffice to remove 
the learned trigger-related features.
BadNets is conducted via poisoning, which means that the parameters of 
all model layers are adjusted during training.
Whereby, to remove the effect of triggers, larger datasets or more epochs 
are required \cite{liu2017Neural}.
Trojaning Attack, in contrast, is a retraining technique that fine-tunes 
the last layers of the models (i.e., it changes less parameters) while 
inserting the trojan (see Section \ref{sec:background_attacks}). 
Hence, one epoch is enough to remove the trigger effect.

At this point, it is important to highlight that
due to the violation of the condition $C3$ as explained Section
\ref{sec:experiments_exp1},  $ConFoc$ produces models with lower values 
in the accuracy with adversarial data than those obtained with benign 
data in the case of Trojaning Attack (WM) (see dark grey cells in the table). 
These values, however are constantly above 90\% and do not tend to decrease 
with the sizes of the healing set.

\textit{\textbf{ConFoc Overhead.}}
There is no clear advantage (with respect to the metrics)
on using either original or transformed inputs with $ConFoc$. 
However, there is a difference in the overhead caused at testing 
time. 
%
%
Transforming the inputs with Algoritm \ref{alg:styled} directly
imposes an 10-run average overhead of 3.14 s with 10 iterations 
of the optimizer LGBFS \cite{liu1989Limited} over a Titan XP GPU.
%
%
%
%
We reduce this runtime overhead to values around 0.015 s 
by applying the principles of Algorithm 1 and Algorithm 3 
to train image transformation neural networks offline 
for each chosen style as proposed in 
\cite{johnson2016PerceptualLoss}. 
This implementation is included in our prototype 
\cite{mvillar2019ConFocRepo}. 
%
%
%
%
%
%
$ConFoc$ does not impose any overhead at testing if original 
inputs are used.

\noindent
\fbox {
  \parbox{8.5cm}{
    \textbf{Findings.} $ConFoc$ outperforms the state-of-the-art method
    regardless the size of the healing set, without imposing any  
    overhead when original inputs are used in the evaluation.  
  }
}

\subsection{Robustness Against Adaptive and Complex Triggers}
\label{sec:experiments_exp6}
    \textbf{RQ6.} How effective is ConFoc protecting DNN models when
    adaptive and complex triggers are used? 

This section 
evaluates $ConFoc$ against trojan attacks conducted with complex triggers.
We conduct the attacks with BadNets because this approach 
extracts trigger-related features in all the layers of the model, 
making it more difficult to eliminate.
The idea is to test of $ConFoc$ in the most complex
scenarios.
We make sure that the attacks comply with the 
conditions $C1$-$C4$ specified in Section 
\ref{sec:background_attack_definition}.
The complex triggers are described below using as reference the trigger
(referred here to as original) and data split presented in Section 
\ref{sec:evaluation_attack1} (see Figure \ref{fig:gtsrb_split}).
%
In the description, the sizes of the triggers correspond to a 
percentage of the larger side of the inputs.

\begin{itemize}

	\item \textbf{Adaptive.} 
	We assume an adaptive attacker knowledgeable about $ConFoc$ 
	who seek to mitigate the healing procedure by infecting the 
	model with styled adversarial samples.
	The original trigger is added to the samples in the trojan set (trj).
	These samples are then transformed via $ConFoc$ using the
	base $b_1 \in B$, which is used in the healing process enacting 
	so the best scenario for the attacker.
	The target class is 19.
	
	\item \textbf{Larger.} 
	A white square of size is 15\% (rather than the 10\% size of original) located 
	in the botton-right corner of the image. 
	The target class is 19. 

	\item \textbf{Random Pixel.} 
	A square of size 10\% located in botton-right corner of the image, 
	whose pixel values are randomly chosen. 
	The target class is 19. 

	\item \textbf{Multiple Marks.} 
	A trigger consisting of four marks: 
	(1) the original white square in the botton-right corner, 
	(2) the random pixel square described above located in the botton-left corner, 
	(3) a white circle (circumscribed by a square of size 15\%) located in 
        top-left corner, and 
    (4) the same circle but filled with random pixels located in the top-right 
        corner.
    The target class is 19.  

	\item \textbf{Many-to-One.}
	Each of the multiple marks described above are added individually to 
	the samples in the trojan set (trj). 
	Namely, we create four trojan sets, each with one of the marks.
	The target class assigned to all the resulting adversarial samples in these 
	sets is 19.

	\item \textbf{Many-to-Many.}
	In this case we assign a different target class to each of trojan sets
	described above.
	The assignment is as follows: 
	(1) botton-right mark targets class 19, 
	(2) botton-left mark targets class 20,
	(3) top-right mark targets class 21, and
	(4) top-left mark targets class 22.
	 
\end{itemize}

\begin{table}[!t]
  \caption{Performance With Adaptive/Complex Triggers}
  \label{tab:complex_triggers}
  \centering
  \scalebox{0.87}{
  \begin{tabular}{lcccccc}
    \toprule
       & \multicolumn{3}{c}{Before ConFoc}
       & \multicolumn{3}{c}{After ConFoc}\\	
    \cmidrule(r){2-4}
    \cmidrule(r){5-7}	
    Trigger 		& Acc (Ben)	&Acc (Adv)	& ASR		& Acc (Ben) &Acc (Adv)	& ASR	    \\		
    \midrule
    Adaptive   		& 98.14\%	& 2.33\%	& 100.0\%   & 98.14\%	& 97.91\%	& 0.00\%	\\	   
    Larger          & 97.67\%	& 2.56\%	& 99.76\%	& 97.67\%	& 97.91\%	& 0.00\%	\\  
    Random Pixel    & 97.91\%	& 2.33\%	& 100.0\%	& 98.14\%	& 97.44\%	& 0.00\%	\\	   	 
    Multiple Marks  & 97.44\%	& 2.33\%	& 100.0\%	& 97.67\%	& 97.91\%	& 0.00\%	\\
    Many-to-One     & 96.51\%	& 20.93\%	& 80.48\%	& 97.44\%	& 97.21\%	& 0.00\%	\\
    Many-to-Many    & 97.91\%	& 21.63\%	& 80.00\%	& 97.91\%	& 98.14\%	& 0.00\%	\\
    \bottomrule
  \end{tabular}
  }
\end{table}

Table \ref{tab:complex_triggers} shows the metric of the trojaned models
before and after applying $ConFoc$. 
Results show that $ConFoc$ effectively reduces the ASR to the minimum while 
ensures both accuracies remain close or better than the initial values.

\noindent
\fbox {
  \parbox{8.5cm}{
    \textbf{Findings.} $ConFoc$ effectively eliminate trojans on DNNs 
    compromised with complex triggers, while ensures accuracy values 
    that in average either equal or outperform the initial values of the model 
    when conditions $C1$-$C4$ are satisfied. 
  }
}

\section{Conclusions and Future Work}\label{sec:conclusion}
We present a generic model hardening technique called $ConFoc$ to 
protect DNNs against trojan attacks. 
$ConFoc$ takes as input an infected model and produces a healed 
version of it. 
These models are healed by fine-tuning them with a small dataset 
of benign inputs augmented with styles extracted from a few random 
images. 
We run experiments on different models and datasets, infected with
a variety of triggers by two different trojan attacks: 
BadNets and Trojaning Attack. 
Results show that $ConFoc$ increasingly reduces the sensitivity 
of trojaned models to triggers as more styles are used in 
the healing process.
We proved that our method can be equally applied to any 
model (trojaned or not) since it does not impact the initial 
accuracy of the model.
In comparison with the state-of-the-art, we validate that
$ConFoc$ consistently correct infected models, regardless the dataset,
architecture or attack variation. 
Our results leads us to new research questions related to the 
internal behavior of models.
Future work will aim to investigate which neurons relate to the content
of inputs. 
This information will be used to devise a novel white-box approach to 
detect misbehaviors based on the activation of these neurons.





%
\bibliographystyle{IEEEtran}
\bibliography{ref}
\end{document}